%% file: main.tex
\documentclass[journal, 10pt]{IEEEtran}
\IEEEoverridecommandlockouts

\usepackage{etoolbox}
\usepackage{capt-of}
\makeatletter
\let\NAT@parse\undefined
\apptocmd\@maketitle{{\eyecatcher{}\par}}{}{}
\makeatother
\usepackage{amsmath,mathtools}
\usepackage{amsfonts}
\usepackage{amssymb}
\usepackage{balance}
\usepackage{graphicx}
\usepackage[skip=2pt,font=small]{caption}
\usepackage{subcaption}
\usepackage[numbers, square, comma,compress]{natbib}
\usepackage{siunitx}
\usepackage{color}
\usepackage[table]{xcolor}
\usepackage{booktabs}
\usepackage{footmisc}
\usepackage{dblfloatfix}
\usepackage{url}
\usepackage{nth}
\usepackage{animate}
\usepackage{comment}
\usepackage{pgfplots}
\usepackage{tikz}
\usetikzlibrary{arrows}
\usepackage[normalem]{ulem}
\usepackage{paralist}
\usepackage[ruled,vlined]{algorithm2e}
\usepackage{booktabs}
\usepackage{siunitx}
\usepackage[utf8]{inputenc}
\usepackage{pgfplots}
\usepackage{multirow}
\usepackage{amsthm}
\usepackage{enumitem}

\DeclareUnicodeCharacter{2212}{−}
\usepgfplotslibrary{groupplots,dateplot}
\usetikzlibrary{patterns,shapes.arrows}
\pgfplotsset{compat=newest}

\newtheorem{proposition}{Proposition}

\newcommand{\rom}[1]{(\expandafter{\romannumeral #1\relax})}
\newcommand{\mat}[1]{\begin{bmatrix}#1\end{bmatrix}}
\newcommand{\bm}[1]{\boldsymbol{#1}}

\definecolor{royalazure}{rgb}{0.0, 0.22, 0.66}
\definecolor{mayablue}{rgb}{0.45, 0.76, 0.98}
\DeclareMathOperator*{\argmin}{argmin}

\definecolor{somegray}{rgb}{0.5, 0.5, 0.5}
\newcommand{\darkgrayed}[1]{\textcolor{somegray}{#1}}

\let\vec\bm
\newcommand{\lb}[1]{\underbar{$#1$}}
\newcommand{\ub}[1]{\overline{#1}}

\newcommand{\change}[1]{#1}
\newcommand{\changeSecond}[1]{#1}
\newcommand{\changeElia}[1]{#1}
\newcommand{\changeThird}[1]{#1}
\newcommand{\changeFourth}[1]{#1}
\newcommand{\changeFifth}[1]{#1}
\newcommand{\changeSix}[1]{#1}

\newcommand{\rebuttal}[1]{#1}
\newcommand{\rebuttaloff}[1]{{\color{red}\sout{}}}
\newcommand{\rebuttaltwo}[1]{#1}
\newcommand{\rebuttalthree}[1]{#1}

\makeatletter
\newcommand*\titleheader[1]{\gdef\@titleheader{#1}}
\AtBeginDocument{%
  \let\st@red@title\@title
  \def\@title{%
    \vskip-2.0em
    \bgroup\normalfont\large\centering\@titleheader\par\egroup
    \vskip0.59em\st@red@title}
}
\makeatother
\titleheader{ \darkgrayed{This paper has been accepted for publication in the
\\ IEEE Transactions on Robotics (T-RO), 2022.
\copyright IEEE} }

\newcommand\eyecatcher{
\centering
\vspace{8pt}
\captionsetup{type=figure}\setcounter{figure}{0}
\includegraphics[width=\linewidth]{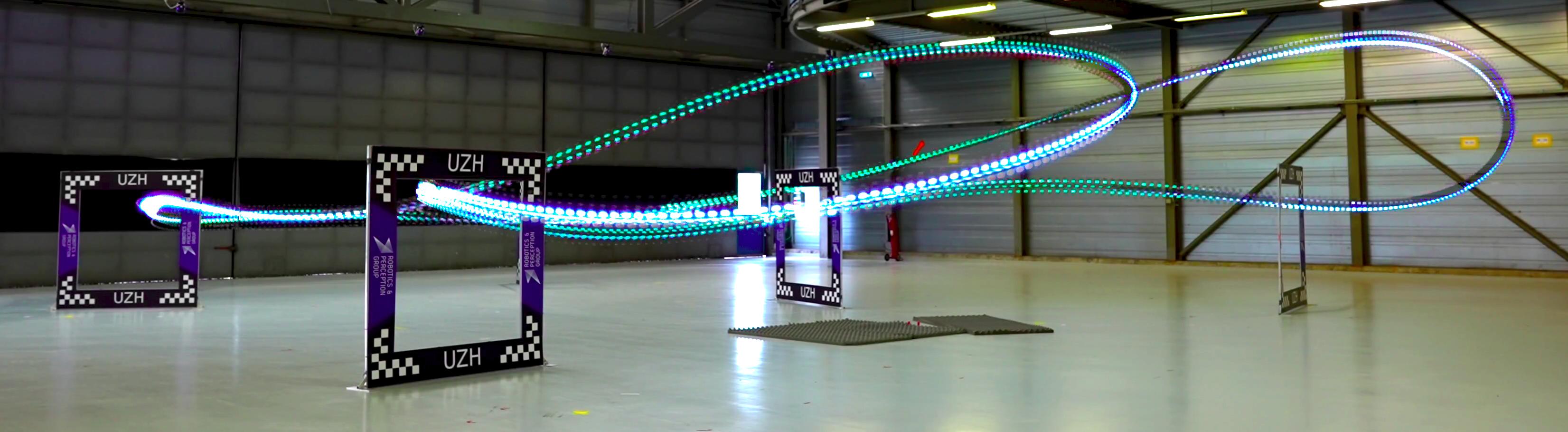}
\captionof{figure}{Execution of a trajectory generated using a point-mass model and tracked by the proposed Model Predictive Contouring Controller. This control approach \change{results} in time-optimal quadrotor flight, achieving velocities of up to \change{\SI[per-mode=symbol]{60}{\kilo\meter\per\hour}}, \changeFifth{accelerations over \SI{5}{\g}}, and lap times that outperform those of world-class professional drone pilots.}
\label{fig:eyecatcher}
\vspace{-20pt}
}
\title{Model Predictive Contouring Control for Time-Optimal Quadrotor Flight}

\author{Angel Romero, Sihao Sun, Philipp Foehn, Davide Scaramuzza
        \thanks{The authors are with the Robotics and Perception Group, University of Zurich, Switzerland (\protect\url{http://rpg.ifi.uzh.ch}).
        This work was supported by the National Centre of Competence in Research (NCCR) Robotics, through the Swiss National Science Foundation (SNSF), and the European Union’s Horizon 2020 Research and Innovation Programme under grant agreement No. 871479 (AERIAL-CORE) and the European Research Council (ERC) under grant agreement No. 864042 (AGILEFLIGHT).
        }
}

\begin{document}

\maketitle

\input{sections/abstract2}

\noindent Video of the experiments: \url{https://youtu.be/mHDQcckqdg4}

\input{sections/introduction}
\input{sections/relatedwork}
\input{sections/problem}

\input{sections/application_quads}

\input{sections/dynamic_allocation}
\input{sections/path_generation}
\input{sections/hover_to_hover}

\input{sections/planning_3D_path}

\input{sections/delay_ablation}
\input{sections/experiments_realworld}

\input{sections/discussion}

{\small
\balance
\bibliographystyle{IEEEtran}
\bibliography{references}
}

\rebuttalthree{
\begin{IEEEbiography}[{\includegraphics[width=1in,height=1.25in,clip,keepaspectratio]{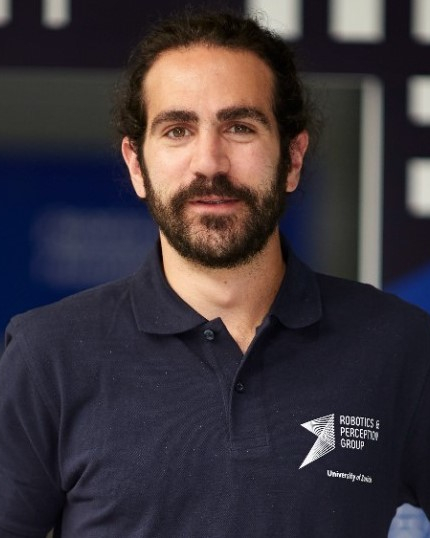}}]{Angel Romero} (1993, Spain) received a MSc degree in "Robotics, Systems and Control" from ETH Zurich in 2018.
Previously, he received a B.Sc. degree in Electronics Engineering from the University of Malaga in 2015.
He is currently working toward a Ph.D. degree in the Robotics and Perception Group at the University of Zurich, finding new limits in the intersection of machine learning, optimal control, and computer vision applied to super agile autonomous quadrotor flight under the supervision of Prof. Davide Scaramuzza.
\end{IEEEbiography}

\begin{IEEEbiography}[{\includegraphics[width=1in,height=1.25in,clip,keepaspectratio]{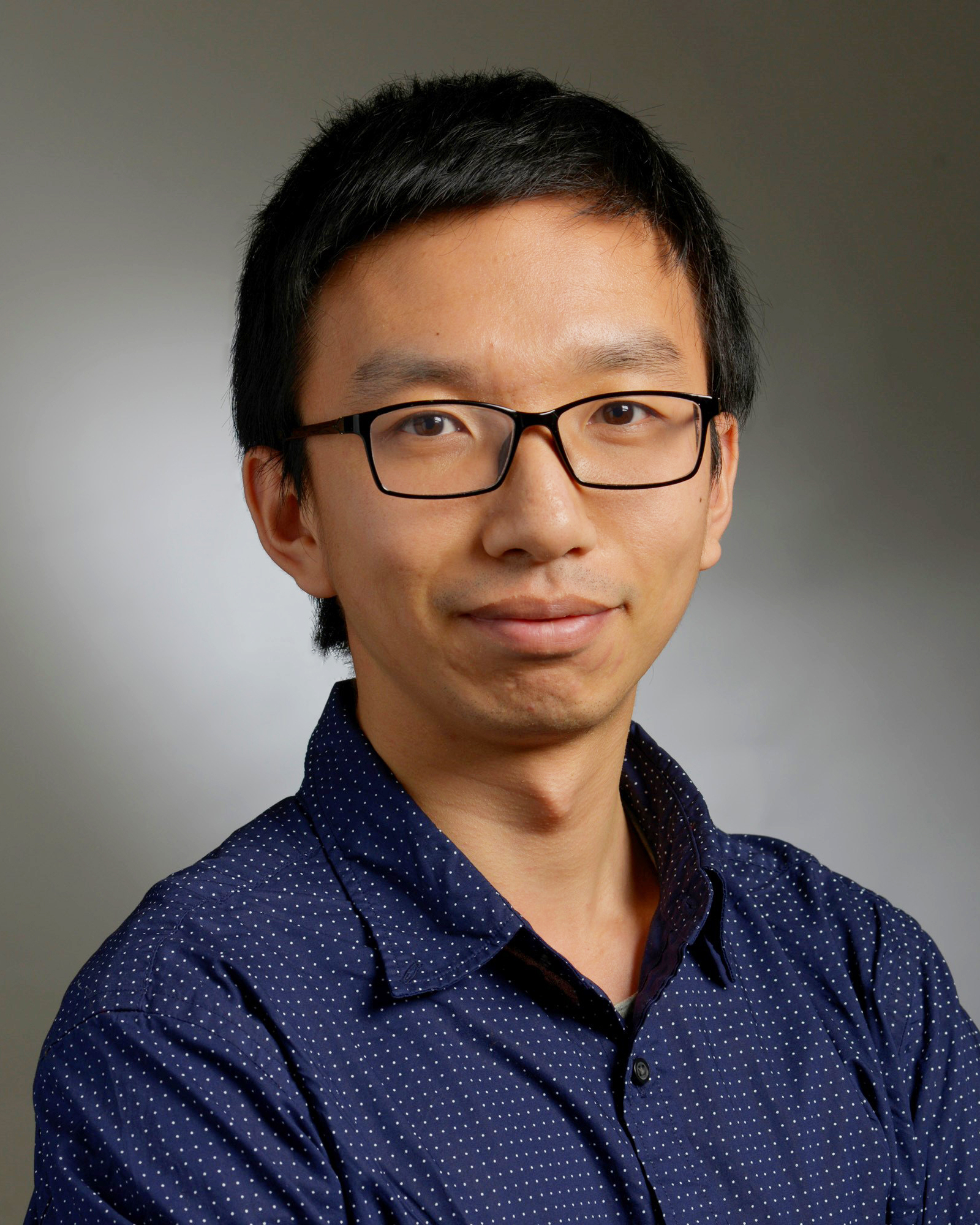}}]{Sihao Sun} (1992, China) received the B.Sc. and M.Sc. degrees in aerospace engineering from Beihang University, Beijing, China, in 2014 and 2017, respectively. In 2020, He received A Ph.D. degree in aerospace engineering from Delft University of Technology, Delft, the Netherlands. From 2020 to 2021, he was first a visiting scholar and then a postdoctoral researcher in the Robotics and Perception Group, University of Zurich, Switzerland. His research interests include system identification, aerial robotics, and nonlinear control.
\end{IEEEbiography}

\begin{IEEEbiography}[{\includegraphics[width=1in,height=1.25in,clip,keepaspectratio]{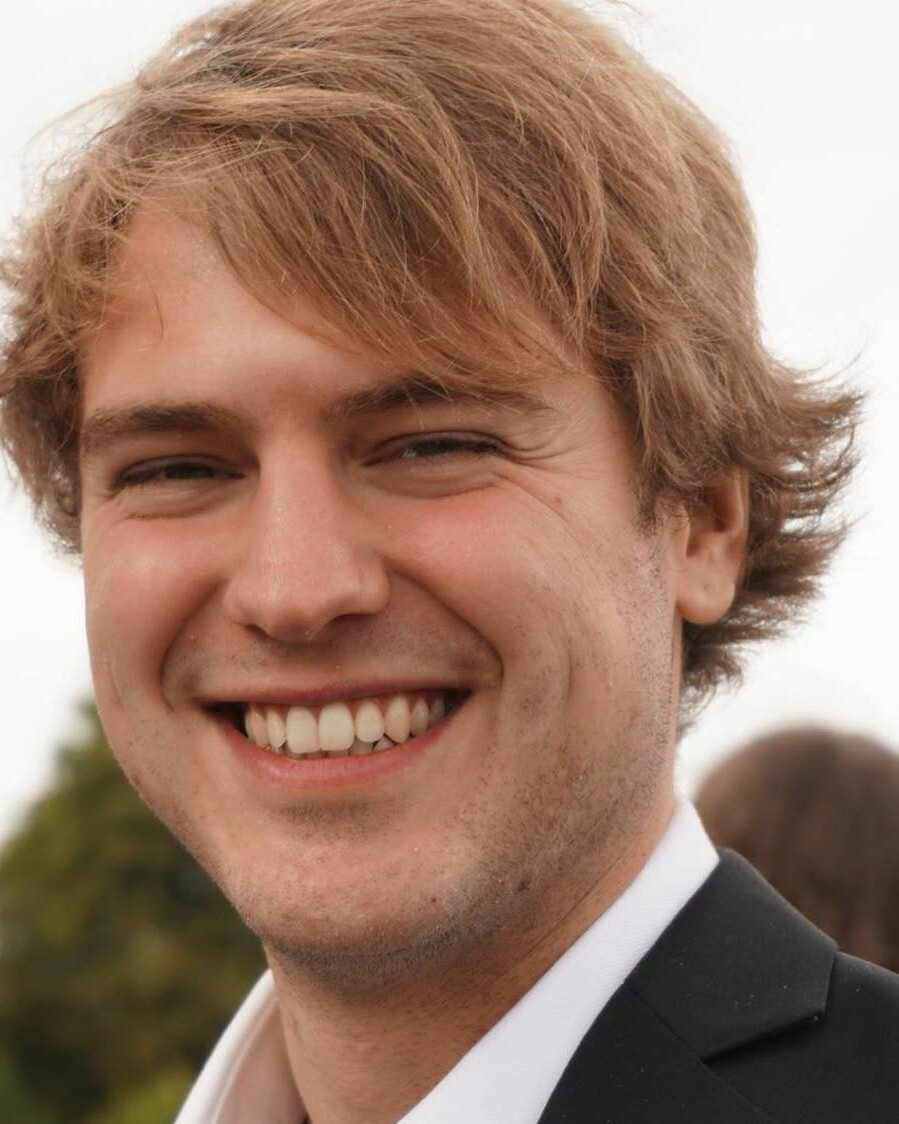}}]{Philipp Foehn} (1991, Switzerland) received a M.Sc. degree in "Robotics, Systems and Control" from ETH Zurich, Switzerland, in 2017.
In 2021, Philipp received his Ph.D. degree in robotics from the University of Zurich, where was working on autonomous agile drones.
His studies focused on trajectory generation, model predictive control, and vision-based state estimation, all in the context of low-latency, agile flight at the limits of the performance envelope of quadrotors.
\end{IEEEbiography}

\begin{IEEEbiography}[{\includegraphics[width=1in,height=1.25in,clip,keepaspectratio]{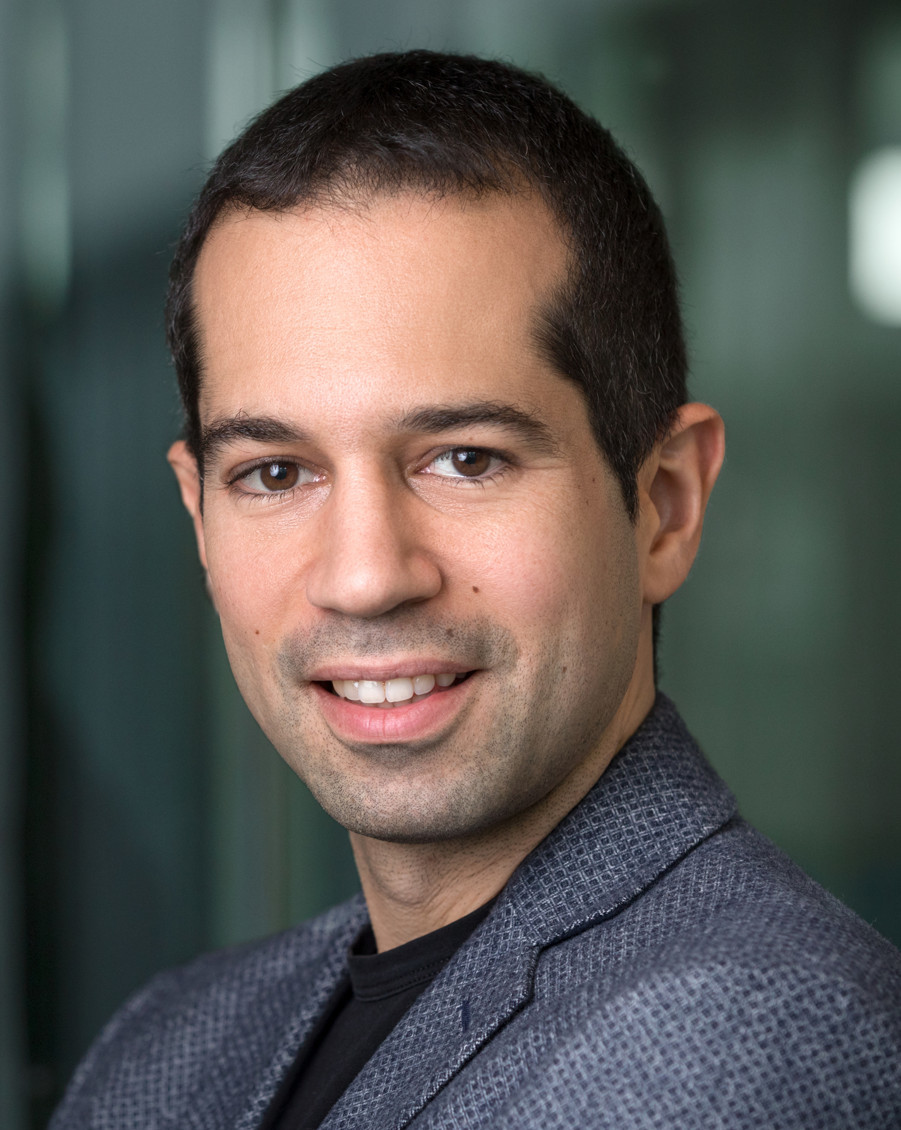}}]{Davide Scaramuzza} (1980, Italy) received a Ph.D. degree in robotics and computer vision from ETH Zurich, Switzerland, in 2008, followed by postdoctoral research at both ETH Zurich at the University of Pennsylvania, Philadelphia, USA.
He is a Professor of Robotics and Perception with the University of Zurich, where he does research at the intersection of robotics, computer vision, and machine learning, aiming to enable autonomous, agile navigation of micro drones using  standard and neuromorphic event-based cameras. From 2009 to 2012, he led the European project sFly, achieving the first autonomous vision-based navigation of microdrones in GPS-denied environments, which inspired the visual-navigation algorithm of the NASA Mars helicopter. He has served as a consultant for the United Nations, including the International Atomic Energy Agency’s Fukushima Action Plan on Nuclear Safety. He coauthored the book Introduction to Autonomous Mobile Robots (MIT Press). For his research contributions, he won prestigious awards, such as a European Research Council (ERC) Consolidator Grant, the IEEE Robotics and Automation Society Early Career Award, an SNF-ERC Starting Grant, a Google Research Award, a Facebook Distinguished Faculty Research Award, and several paper awards. In 2015, he co-founded Zurich-Eye, today Facebook Zurich, which developed the hardware and software tracking modules of the Oculus VR headset, which sold over 10 million units. Many aspects of his research have been prominently featured in wider media, such as The New York Times, The Economist, Forbes, BBC News, Discovery Channel.
\end{IEEEbiography}
}

\end{document}

%% file: sections/abstract2.tex
\begin{abstract}

We tackle the problem of flying time-optimal trajectories through multiple waypoints with quadrotors.
State-of-the-art solutions split the problem into a planning task---where a global, time-optimal  trajectory is generated---and a control task---where this trajectory is accurately tracked. However, at the current state, generating a time-optimal trajectory that \rebuttal{considers the full quadrotor model requires solving a difficult time allocation problem via optimization, }which is computationally demanding (in the order of minutes or even hours). This is detrimental for replanning in presence of disturbances. 
%
We overcome this issue by \rebuttal{solving the time allocation problem and the control problem} concurrently via Model Predictive Contouring Control (MPCC). Our MPCC optimally selects the future states of the platform at runtime, while maximizing the progress along the reference path and minimizing the distance to it. 
We  show  that, even  when tracking  simplified  trajectories,  the  proposed MPCC  results in  a  path  that  approaches the  true  time-optimal  one, and which  can  be  generated  in  real-time.  
We validate our approach in  the real world, where we  show  that  our  method  outperforms both the current state-of-the-art and a world-class human pilot in terms  of  lap  time  achieving  speeds  of  up  to  60 km/h.
\end{abstract}

%% file: sections/introduction.tex
\section{Introduction}

Multirotor drones are extremely agile, so agile, in fact, that they have become an essential tool for time-critical missions, such as search and rescue, aerial delivery, and even flying cars~\cite{Loianno2020jfr,air_taxi_2020}. For this reason, over the past decade, research on autonomous multirotor drones has continually pushed platforms to higher speeds and agility~\cite{Mellinger12ijrr,loianno2016estimation,kaufmann18icra,Mohta18jfr,Zhou19TRO,Foehn20rss,croonRAS20,MPCSurveyASL_2020,DDAKaufmann_Loquercio_Scaramuzza_2020,foehn2021CPC}. Several competitions have been organized, such as the Autonomous Drone Racing series at the recent IROS and NeurIPS conferences~\cite{Moon19jirc,cocoma2019towards,Madaan20arxiv} and the AlphaPilot challenge~\cite{Foehn20rss,guerra2019flightgoggles}. Their goal: to develop autonomous systems that will eventually outperform expert human pilots. Million-dollar projects, such as AgileFlight~\cite{AgileFlight} and Fast Light Autonomy (FLA)~\cite{Mohta18jfr}, have also been funded by the European Research Council and the United States government, respectively, to further advance the field.

In this paper, we address the problem of flying a quadrotor through multiple waypoints in minimum time. This is often desired for delivery and inspection tasks, and, in the context of search and rescue and drone racing, it is even the ultimate goal. Professional drone racing pilots achieve this with astonishing performance, flying their quadrotors through race tracks at speeds difficult to achieve for autonomous systems. 

Time-optimal multi-waypoint flight raises fundamental challenges for robotics research in terms of planning, control, aerodynamics, and modeling. These become increasingly complex in the presence of high accelerations and aggressive attitude changes. Exploring new ways of generating and tracking minimum-time trajectories not only addresses these challenges, but also has direct implications in what is physically possible for a quadrotor given its limited actuation, ultimately pushing the frontiers of current and emerging real-world applications.

State-of-the-art approaches for time-optimal multi-waypoint flight~\cite{foehn2021CPC, blakbox_karaman_2020} split this problem into a planning task---where a global, time-optimal  trajectory is generated---and a control task---where this trajectory is accurately tracked. 

While the control task can be solved in real-time, at the current state of the art, generating a feasible, global, time-optimal trajectory that considers the full quadrotor dynamics, including single-rotor thrust constraints, \rebuttal{requires solving a very complex time allocation problem, which is} computationally demanding (in the order of minutes or even hours)~\cite{foehn2021CPC}.
This is detrimental for replanning in presence of disturbances. 
When the platform is at the limit of actuation, the slightest deviation from the pre-planned trajectory may result in a sub-optimal flight path or even in a crash. To circumvent this, the trajectory needs to be either planned with conservative actuation limits---which sacrifices speed (\cite{foehn2021CPC}, Section IV.E)---or be replanned online---which is not possible with the current solver times.
Thereby, all computationally efficient time-optimal waypoint planners resort to either modeling the platform as a point-mass or approximating trajectories with polynomials.  
For point-mass model approaches, the problem of finding time-optimal, point-to-point trajectories has a closed form solution \cite{Maurer} and is, therefore, very fast to solve. 
%
This low computational burden has enabled a number of sampling-based methods for quadrotors,
including time-optimal, multi-waypoint flight \cite{Foehn20rss}.
However, these simplified trajectories lack the notion of 3D rotation and are dynamically infeasible (since quadrotors are underactuated systems, they need to rotate to align their thrust with the desired acceleration direction).
On the other hand, polynomial trajectories offer a fast way of generating feasible paths. However, polynomial control inputs are smooth and cannot fully exploit the actuator potential, rendering control policies sub-optimal.

We show that these limitations can be overcome by changing the paradigm and using a Contouring Control approach~\cite{Lam_Manzie_Good_2010, Liniger_Domahidi_Morari_2015} instead of standard trajectory tracking methods. 

\subsection*{Contribution}
In this paper, we propose an MPCC method that considers the full quadrotor dynamics and the real single-rotor thrust constraints to achieve time-optimal quadrotor flight.
In contrast to standard trajectory tracking control methods, our MPCC approach balances the maximization of the progress along a given nominal path and the minimization of the distance to it. \rebuttal{This nominal path can be any continously differentiable 3D path, and does not need to be parameterized by time.}
%
Instead of relying on a pre-computed, computationally expensive \rebuttal{trajectory, the proposed MPCC solves the time allocation problem online}, and has the freedom to optimally select at runtime the states at which the reference trajectory is sampled such that the progress is maximized and the platform stays within actuator bounds.
Furthermore, we demonstrate that this control approach is particularly well suited for \change{the problem of} autonomous drone racing.
\changeFourth{To accurately pass through the waypoints, we encode} their positions within our cost function by dynamically allocating the contour weight. 
Since the reference 3D path does not need to be feasible, we efficiently generate it by using a \change{simpler} point-mass model.
The proposed controller renders speeds, accelerations, and lap times very similar to the theoretical time-optimal trajectories \cite{foehn2021CPC} while keeping the computational burden low enough to be solved in real-time.

\changeFourth{We show that the proposed MPCC can make better use of the available actuator potential and results in faster real-flight lap times than both a standard MPC controller tracking a time-optimal trajectory and a world-class professional human pilot.}

%% file: sections/relatedwork.tex
\section{Related Work}


\subsection{Quadrotor Trajectory Tracking Control}

Nonlinear controllers have been proposed to address quadrotors' complex attitude dynamics, such as the geometric controller~\cite{lee2010geometric}, quaternion-based controller ~\cite{fresk2013full}, tilt-prioritized controller~\cite{brescianini2018tilt}. Model Predictive Control (MPC) is another trajectory tracking solution that simultaneously handles nonlinearities and input constraints~\cite{Bangura14ifac, Diehl2006springer}. Aerodynamic effects are detrimental to tracking accuracy, especially in high-speed flights. Various methods are proposed to tackle aerodynamic effects, such as using first-principle models~\cite{svacha2017improving}, Gaussian-process models~\cite{torrente2021data}, or leveraging accelerometer measurements~\cite{tal2020accurate}. All these control strategies rely on a previously computed dynamically feasible sequence of states and inputs to track. The problem of generating this sequence is called trajectory planning.

\subsection{Quadrotor Time-Optimal Trajectory Planning}
Quadrotor trajectory planning has been extensively studied in the literature. Polynomial trajectories are the first category of planning algorithms, such as the now widely used minimum-snap trajectories~\cite{Mellinger11icra,Mellinger12ijrr}. Given a 4\textsuperscript{th} order smooth trajectory, the full state at each point on the trajectory can be derived using the differential flatness property of the quadrotor.

Autonomous drone racing considers the problem of traversing a path in minimum time while passing through the designated waypoints (gates). \changeElia{To this end, polynomial-based multi-waypoint trajectory planning algorithms have been widely used in the literature due to their simplicity.} However, \changeElia{in the context of completing a race track in minimum time,} sampling states and inputs that are inherently polynomials result in sub-optimal policies. Due to their continuity and smoothness, polynomial trajectories are not able to render control inputs that constantly maximize accelerations during a certain time segment (an issue already observed in~\cite{foehn2021CPC}).
By contrast, optimization-based trajectory planning allows to independently select the optimal sequence of states and inputs at every time step, which inherently considers time minimization while complying with quadrotor dynamics and input constraints. 
Optimization-based approaches have been extensively considered in the literature, ranging from exploiting point-mass models~\cite{Foehn17rss}, simplified quadrotor models~\cite{Hehn12ar, Loock13ecc}, and full-state quadrotor models~\cite{foehn2021CPC, spedicato2017minimum}.


Regarding sampling-based methods, several time-minimizing approaches have been suggested. In \cite{Foehn20rss, Allen16gnc}, a point-mass model is used for the high-level time-optimal trajectory planning. In \cite{Allen16gnc}, an additional trajectory smoothing step is done where they connect the generated trajectory with high order polynomials by leveraging the differential flatness property of the quadrotor. 
Similarly, in~\cite{sampling_lq_mt_Kumar_2017, sampling_SE3_Kumar_2018} the authors take advantage of the differential flatness property in order to search in the space of polynomial, smooth minimum-jerk trajectories while also including the minimization of the time in the cost function. 
However, these approaches need to relax the single actuator constraints and \changeElia{instead limit} the per-axis acceleration, which \changeElia{results in} control policies that are conservative and sub-optimal given a minimum time objective. 
Furthermore, since the authors use polynomials, these sampling approaches can only generate smooth control inputs, meaning that they cannot, for instance, continuously apply full thrust if required. 
The MPCC approach proposed in our work circumvents these problems by separating the planning, which is done using a point-mass model, from the time-optimal objective, which is ultimately handled by the controller. This allows for non-smooth control inputs that respect the true actuator constraints to take full advantage of the input authority of the platform.

Apart from time-optimality, complying with intermediate waypoint constraints is another requirement for path planning in autonomous drone racing. 
A common practice of solving a trajectory optimization problem with waypoint constraints is allocating waypoints to specific time steps and minimizing the spatial distance between these waypoints and the position at the corresponding allocated time steps on the reference trajectory (e.g., ~\cite{jorris2009three, bousson20134d}). 
The time allocation of the waypoints is, however, non-trivial and difficult to determine. 
This is tackled in~\cite{spedicato2017minimum} which, however, uses body rates and collective thrust as control inputs and does not represent realistic actuator saturation.
Recent work~\cite{foehn2021CPC} introduces a complementary progress constraints (CPC) approach, which considers true actuator saturation, uses single rotor thrusts as control inputs, and exploits quaternions \changeElia{to allow full, singularity-free representation of the orientation space with consistent linearization characteristics.}
While the above methods guarantee global optimality of the quadrotor trajectory passing through gates, they are computationally costly and hence intractable in real-time.

\subsection{Model Predictive Contouring Control}


In contrast to solving trajectory planning and control separately, Model Predictive Contouring Control (MPCC) combines trajectory generation and tracking into a single optimization problem~\cite{faulwasser2009model, Lam_Manzie_Good_2010}.
This method minimizes the distance to the reference while maximizing the progress along it in a receding horizon fashion, thereby giving more freedom to the controller to determine its state trajectory. 
\changeElia{The MPCC method only needs a continuously differentiable 3D path as reference, hence dropping the feasibility requirement that standard MPC methods have.}
Furthermore, it allows adding additional constraints and cost functions for obstacle avoidance, such as in~\cite{Schwarting_Alonso-Mora_Paull_Karaman_Rus_2018, Ji_Zhou_Xu_Gao_2021,brito2019model}.
Most pieces of literature apply MPCC on biaxial systems, such as CNC machines~\cite{tang2011predictive}, ground vehicles~\cite{Liniger_Domahidi_Morari_2015,Schwarting_Alonso-Mora_Paull_Karaman_Rus_2018}, and autonomous ground robots~\cite{brito2019model}. In \cite{Liniger_Domahidi_Morari_2015}, the authors apply MPCC in the context of car racing and discuss that the latter offers control policies that are closer to human-like driving than the ones from standard MPC methods. In~\cite{Ji_Zhou_Xu_Gao_2021} the corridor-based MPCC is proposed for drone flights. A simplified 3\textsuperscript{rd} order linear integral model is adopted where jerk is regarded as the control input. However, the nonlinear dynamics and the input constraints are neglected, limiting the method to relatively low-speed flights.

In this paper, we propose an MPCC architecture that considers the more challenging three-dimensional space for autonomous drone racing.
To fully exploit the capability and realize optimal flight performance, we formulate the optimization problem using the full nonlinear quadrotor model.
Additionally, a novel cost function is designed to guarantee the traversal of gates.


%% file: sections/problem.tex
\section{Methodology}
\label{sec:problem_formulation}
Consider the discrete-time dynamic system of a quadrotor with continuous state and input spaces, $\vec{x}_k \in \mathcal{X}$ and $\vec{u}_k \in \mathcal{U}$ respectively. 
Let us denote the time discretized evolution of the system \mbox{$f : \mathcal{X} \times \mathcal{U} \mapsto \mathcal{X}$} such that
\begin{align}
\vec{x}_{k + 1} = f(\vec{x}_k, \vec{u}_k) \notag
\end{align}
where the sub-index $k$ is used to denote states and inputs at time $t_k$.

The general Optimal Control Problem considers the \changeElia{task} of finding a control policy $\pi(\vec{x})$, a map from the current state to the optimal input, $\pi : \mathcal{X} \mapsto \mathcal{U}$, such that the cost function $J: \mathcal{X} \mapsto \mathbb{R}^+$ is minimized:

\begin{align}
\pi(\vec{x}) = & \argmin_u
  & & 
    J(\vec{x}) \notag\\
        & \text{subject to} && \vec{x}_0 = \vec{x} \notag \\ 
        &&& \vec{x}_{k+1} = f(\vec{x}_k, \vec{u}_k) \notag\\
        &&& \vec{x}_k \in \mathcal{X}, \quad \vec{u}_k \in \mathcal{U}
        \label{eq:ocp}
\end{align}

\subsection{Model Predictive Control}
For standard Model Predictive Control approaches \cite{MPCSurveyASL_2020, Bangura14ifac, Diehl2006springer}, the objective is to minimize a quadratic penalty on the error between the predicted states and inputs, and a given dynamically feasible reference $\vec{x}_{k,ref}$ and $\vec{u}_{k, ref}$. 
Consequently, the cost function $J(\vec{x})$ in problem \eqref{eq:ocp} is substituted by:
\begin{align}
    \label{eq:J_mpc}
    J_{MPC}(\vec{x}) = \sum_{k = 0}^{N-1} \Vert \Delta \vec{x}_k \Vert_{\bm{Q}}^2 + \Vert \Delta \vec{u}_k \Vert_{\bm{R}}^2 + \Vert \Delta \vec{x}_N \Vert_{\bm{P}}^2
\end{align}
where $\Delta \vec{x}_k = \vec{x}_k - \vec{x}_{k,ref}$, $\Delta \vec{u}_k = \vec{u}_k - \vec{u}_{k,ref}$, and where $\vec{Q} \succeq 0$, $\vec{R} \succ 0$ and $\vec{P} \succeq 0$ are the state, input and final state weighting matrices. 
The norms of the form $\Vert \cdot \Vert_A^2$ represent the weighted Euclidean inner product $\Vert \vec{v} \Vert^2_A = \vec{v}^T A \vec{v}$. 

This formulation relies on the fact that a feasible reference $\vec{x}_{k,ref}$, $\vec{u}_{k,ref}$ is accessible for the future time horizon $N$. 
Searching for these is often referred to as \emph{planning}, and for several applications, it becomes rather arduous and computationally expensive \cite{foehn2021CPC, Song_Steinweg_Kaufmann_Scaramuzza_2021}.
\subsection{Model Predictive Contouring Control}
\changeElia{In contrast to standard MPC methods, where a time-sampled state and input reference is tracked, MPCC methods consider the higher-level task of minimizing the Euclidean distance to a three-dimensional path while maximizing the speed at which the path is traversed.}

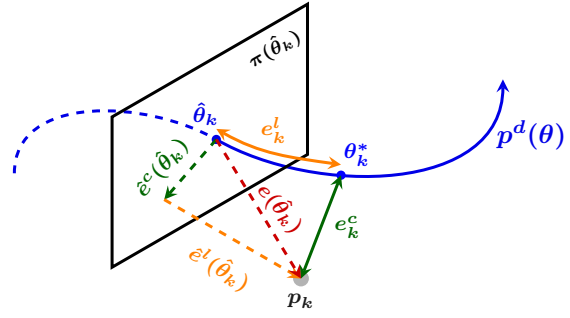
\begin{figure}[t]
  \centering
  \centering
  \begin{subfigure}{\linewidth}
    \input{tikz/tikz_3D_errors_explanation}
    \vspace{0.5cm}
  \end{subfigure}
  \caption{\rebuttaltwo{Since $e^l_k$ and $e^c_k$ cannot be easily computed, they need to be approximated. Given the current position of the platform $\vec{p}_k$, the position error is written as $\vec{e}(\hat{\theta}_k) = \vec{p}_k - \vec{p}^d(\hat{\theta}_k)$, shown in dashed red. When $\vec{e}(\hat{\theta}_k)$ is projected onto the plane $\pi(\hat{\theta}_k)$, it results in $\vec{\hat{e}}^c(\hat{\theta}_k)$ (shown in dashed green), which is the approximation to the contour error $e_k^c$. Similarly, the projection of $\vec{e}(\hat{\theta}_k)$ onto the normal direction of $\pi(\hat{\theta}_k)$ results in $\hat{\vec{e}}^l(\hat{\theta}_k)$, which is the approximation to the lag error $e^l_k$.} Note that, graphically, if $\hat{\vec{e}}^l(\hat{\theta}_k)$ is zero, then $e^l_k$ is also zero, and therefore $e^c_k$ is equal to $\Vert \hat{\vec{e}}^c(\hat{\theta}_k) \Vert$ and $\theta^*_k$ is equal to $\hat{\theta_k}$.}
\label{fig:contour_lag_errors_explanation}
\end{figure}

\rebuttal{Let us denote the arc length of the reference path (or \emph{progress}) as $\theta$, and the same arc length at timestep $k$ as $\theta_k$.}
Let us also denote the desired arc length parameterized 3D path as \rebuttal{$\vec{p}^d(\theta) = \begin{bmatrix}x^d(\theta) & y^d(\theta) & z^d(\theta)\end{bmatrix}^T$}, and the system's position \rebuttal{at time step $k$} as $\vec{p}_k = \begin{bmatrix}x_k & y_k & z_k\end{bmatrix}^T$. 
The Model Predictive Contouring Control formulation (MPCC \cite{Lam_Manzie_Good_2010}) attempts to minimize the projected distance from the current position $\vec{p}_k$ to the desired position $\vec{p}^d(\theta)$ while maximizing the progress $\theta$ along it. 
This is referred to in the literature as the \emph{Contouring Control Problem} \cite{Lam_Manzie_Good_2010}, and it can be written in the same shape as problem \eqref{eq:ocp} by substituting the cost function by:
\rebuttal{
\begin{align}
    \label{eq:J_mpcc}
    J_{MPCC}(\vec{x}) = \sum_{k = 0}^{N} q_c (e^c_{k})^{2} - \rho \theta_N
\end{align}
}
where $e^c_k \in \mathbb{R}^+$ is the \emph{contour error} \rebuttal{(solid green line in Fig. \ref{fig:contour_lag_errors_explanation})} at step $k$ and is defined as the projection of $\vec{p}_k$ over $\vec{p}^d(\theta_k)$,
\rebuttal{
\begin{align}
\label{eq:contour_error_def}
    e^{c}_k = \min_{\theta} \quad \Vert \vec{p}_k - \vec{p}^d(\theta)\Vert_2
\end{align}
where the minimizer at time step $k$ is depicted in Fig. \ref{fig:contour_lag_errors_explanation} as $\theta^*_k$.
In cost function \eqref{eq:J_mpcc}, the negative sign on $\rho \in \mathbb{R}^+$ indicates that the progress at the last step, $\theta_N$, is maximized. 
Finally, $q_c \in \mathbb{R}^+$ is the \emph{contour weight}.

However, this formulation of the contour error is not well suited to be used inside an online optimization scheme because it is an optimization problem by itself \eqref{eq:contour_error_def}. 
To overcome this limitation, let us introduce $\hat{\theta}_k$, an approximation to $\theta_k^*$ with dynamics
\begin{align}
    \hat{\theta}_{k + 1} = \hat{\theta}_k + v_{\hat{\theta}} \Delta t
    \label{eq:theta_dyn}
\end{align}
where $v_{\hat{\theta}} = \frac{\Delta \hat{\theta}_k}{ \Delta t}$ is a virtual control that will be determined by the controller at every timestep.
Let us also denote $\pi(\theta_k)$ as the plane that is normal to the curve $\vec{p}^d(\theta_k)$ at point $\theta_k$, 
\rebuttaltwo{and let the position error at time $k$ be $\vec{e}(\hat{\theta}_k) = \vec{p}_k - \vec{p}^d(\hat{\theta}_k)$ (shown in dashed red in Fig. \ref{fig:contour_lag_errors_explanation}).
The contour error $e^c_k$ can then be approximated by the norm of the projection of $\vec{e}(\hat{\theta}_k)$ onto the normal plane $\pi(\hat{\theta}_k)$.} This projection is defined as $\hat{\vec{e}}^c(\hat{\theta}_k)$ (dashed green line in Fig. \ref{fig:contour_lag_errors_explanation}).
The arc length corresponding to the minimizer of \eqref{eq:contour_error_def}, i.e., $\theta^*_k$, and the arc length of its approximation, i.e., $\hat{\theta}_k$, are linked by the \emph{lag error} $e^l_k = \Vert \theta^*_k - \hat{\theta}_k\Vert$ (solid orange line in Fig. \ref{fig:contour_lag_errors_explanation}).
However, since $\theta^*_k$ is not known at optimization time, the true lag error $e^l_k$ is also approximated by the norm of $\hat{\vec{e}}^l(\hat{\theta}_k)$ (dashed orange line in Fig. \ref{fig:contour_lag_errors_explanation}), \rebuttaltwo{which is defined as projection of $\vec{e}(\hat{\theta}_k)$ onto the normal direction of $\pi(\hat{\theta}_k)$.}
}

\begin{proposition}
If \rebuttal{$\hat{\vec{e}}^l(\hat{\theta}_k) = \vec{0}$}, the approximations \rebuttal{$\hat{\theta}_k$, $\hat{\vec{e}}^l(\hat{\theta}_k)$,} and \rebuttal{$\hat{\vec{e}}^c(\hat{\theta}_k)$} are equal to their respective true values.
\end{proposition}

\rebuttal{
\begin{proof}
$e^c_k$ is defined in \eqref{eq:contour_error_def} as the minimum distance between $p_k$ and $p^d(\theta)$, and therefore, $p_k \subset \pi(\theta^*_k)$.
If $\hat{\vec{e}}^l(\hat{\theta}_k) = \vec{0}$, then $\vec{p}_k \subset \pi(\hat{\theta}_k)$ as well.
It follows that $\pi(\hat{\theta}_k) \equiv \pi(\theta^*_k)$. Therefore, $\hat{\theta}_k = \theta^*_k$, $e^c_k = \Vert \hat{\vec{e}}^c(\hat{\theta}_k) \Vert$ and $e^l_k = \Vert \hat{\vec{e}}^l(\hat{\theta}_k) \Vert = 0$. 
\end{proof}
}
Consequently, to ensure that these approximations are accurate it is necessary to include the minimization of \rebuttal{$\Vert \hat{\vec{e}}^l(\hat{\theta}_k) \Vert$} in the cost function of problem \eqref{eq:ocp_generic}.
\rebuttaloff{Now, consider $v_{\hat{\theta}} = \frac{d \hat{\theta}}{ d t}$, the speed at which $\hat{\theta}_k$ evolves, discretized, as $\hat{\theta}_{k + 1} = \hat{\theta}_k + v_{\hat{\theta}} \Delta t$. Hence, we can write.}
\rebuttal{Now, considering \eqref{eq:theta_dyn}, we can write $\hat{\theta}_N$ as}
\begin{align}
  \hat{\theta}_N = \hat{\theta}_0 + \sum_{k=0}^N v_{\hat{\theta},k} \Delta t
  \label{eq:theta_N}
\end{align}

Adding \rebuttal{$\Vert \hat{\vec{e}}^l(\hat{\theta}_k) \Vert$} and \eqref{eq:theta_N} to \eqref{eq:J_mpcc}, the optimization in problem \eqref{eq:ocp} becomes:
\rebuttal{
\begin{align}
  \label{eq:ocp_generic}
  \pi(\vec{x}) = & \argmin_u
  & & \sum_{k = 0}^{N} \Vert 
  \hat{\vec{e}}^c(\hat{\theta}_k) \Vert_{q_c}^2 + \Vert \hat{\vec{e}}^l(\hat{\theta}_k) \Vert_{q_l}^2 - \mu v_{\hat{\theta},k} \notag\\
        & \text{subject to} & & \vec{x}_0 = \vec{x} \notag \\
        &&& \vec{x}_{k+1} = f(x_k, u_k) \notag \\
        &&& \vec{x}_k \in \mathcal{X} \quad \vec{u}_k \in \mathcal{U}
\end{align}
}
where $q_l \in \mathbb{R}^+$ is the \emph{lag weight}, the maximization of $\theta_N$ has been rewritten in terms of $v_{\theta,k}$ and $\mu \in \mathbb{R}^+$ is the new progress weight variable.
\rebuttal{To ensure an accurate progress approximation, the weight on the lag error $q_l$ is chosen high as suggested in \cite{Lam_Manzie_Good_2010, Liniger_Domahidi_Morari_2015}}.
From this point on, we drop the hat nomenclature for the approximations of the errors for readability. 
Thus, $\theta_k$, \rebuttal{$\vec{e}^c(\theta_k)$, $\vec{e}^l(\theta_k)$} and $v_{\theta}$ directly refer to their corresponding approximations.

This control strategy offers multiple advantages that are exploited in this paper. 
Particularly, since the maximization of the progress is directly encoded in the optimization problem being solved online, it benefits from the freedom to select the best times at which the states along the reference 3D path are sampled. 
This way, the MPCC formulation eliminates the need for a set of pre-sampled states and inputs. Thus the only requirement is an arc length parametrized, continuously differentiable 3D path to track, $\vec{p}^d(\theta)$.

\subsection{Arc Length Parameterization of the Paths}
\label{sec:arc_length_splines}
Arc length parameterization of general curves is a non-trivial problem, and finding a closed-form solution for an arc-length parameterized curve given another arbitrary parameterization is, in general, not possible. 
Instead, we find approximations to arc length parameterized paths. 
Let $\vec{p}^d(t)$ be the 3D path to track, and it can be given in two forms: as a continuous time-dependent curve or as a sequence of sampled points in time. 
If it is given as a continuous curve, we use the \emph{bisection method} from \cite{ArcLengthWang2002} where the curve is sampled such that the arc length between samples is constant by numerically computing the integral and doing binary search until convergence. 
If it's given as a sequence of sampled points, we search for these equidistant segments by assuming linearity between samples. 
In both cases, for each point, we store the corresponding arc length, the position, and the normalized velocity. 
This process results in a sequence of $P$ of these points to which we fit 3\textsuperscript{rd} order splines $\vec{\rho}_i(\theta_k)\; i \in [0, \dots, P-1]$ as shown in \eqref{eq:arc_length}. 
For further details, we refer the reader to \cite{ArcLengthWang2002}.

\begin{align}
  \vec{p}^d(\theta_k) =
  \begin{cases}
    \vec{\rho}_0(\theta_k) & 0 \leq \theta_k \leq \theta_0 \\
    \vec{\rho}_1(\theta_k) & \theta_1 \leq \theta_k \leq \theta_2 \\
    \hspace{0.5cm}\vdots \\
    \vec{\rho}_{P-1}(\theta_k) & \theta_{P-1} \leq \theta_k \leq \theta_{P}
  \end{cases}
  \label{eq:arc_length}
\end{align}

\subsection{Derivation of Contour and Lag Errors in 3D}
\label{sec:lag_contour_errors}
\rebuttal{
Let us define $\vec{t}(\theta_k) \in \mathbb{R}^3$, the tangent line of $\vec{p}^d(\theta_k)$ evaluated at $\theta_k$. From the definition of arc-length,
\begin{align}
\theta_k &= \int_0^{\theta_k}\Vert (\vec{p}^d)'(t)\Vert dt \iff \notag \\
\frac{d\theta_k}{d\theta_k} &= \frac{d \int_0^{\theta_k}\Vert (\vec{p}^d)'(t)\Vert dt}{d \theta_k} \iff \label{eq:arc_length_norm} \\
1 &= \Vert (\vec{p}^d)' (\theta_k)\Vert = \left\lVert \frac{d \vec{p}^d(\theta_k)}{d \theta_k} \right\rVert = \Vert \vec{t}(\theta_k) \Vert \notag
\end{align}
where in the first equivalence we have differentiated both sides of the equation and the second equivalence follows from the Fundamental Theorem of Calculus.
}
\rebuttaloff{Consider the tangent line of the desired path, $\vec{t}(\theta_k) \in \mathbb{R}^3$, as shown in Fig. \ref{fig:contour_lag_errors_explanation} (bottom),}

$\vec{e}(\theta_k)$ can be decomposed in a component that is projected onto $\vec{t}(\theta_k)$, $\vec{e}^l(\theta_k)$, and a component contained in $\pi(\theta_k)$, which is $\vec{e}^c(\theta_k)$, such that $\vec{e}(\theta_k) = \vec{e}^l(\theta_k) + \vec{e}^c(\theta_k)$.
Now, the expression for the projection of $\vec{e}(\theta_k)$ onto $\vec{t}(\theta_k)$ is:
\begin{align}
  \rebuttal{\vec{e}^l(\theta_k)} = \left(\vec{t}(\theta_k)^T \cdot \vec{e}(\theta_k) \right) \vec{t}(\theta_k) = \rebuttal{e^l(\theta_k)} \vec{t}(\theta_k) \notag
\end{align}
We are interested in minimizing the $q_l$-weighted Euclidean norm of this error,
\begin{align}
\rebuttal{\Vert\vec{e}^l(\theta_k)\Vert_{q_l}^2} &= \rebuttal{\Vert e^l(\theta_k) \Vert_{q_l}^2} \cdot \Vert \vec{t}(\theta_k)\Vert_{q_l}^2 \notag\\
&= q_l\rebuttal{e^l(\theta_k)^2}
\label{eq:ql_norm}
\end{align}
\rebuttal{which follows by the result from \eqref{eq:arc_length_norm}.}
Similarly, we can write the contour error as
\begin{align}
    \begin{multlined}
      \rebuttal{\vec{e}^c(\theta_k)} = \vec{e}(\theta_k) - \rebuttal{\vec{e}^l(\theta_k)} = \notag\\
      \begin{bmatrix}
        1-t^2_x(\theta_k) & -t_x(\theta_k) t_y(\theta_k) & -t_x(\theta_k) t_z(\theta_k)\\
        -t_x(\theta_k) t_y(\theta_k) & 1-t_y^2(\theta_k) & -t_y(\theta_k) t_z(\theta_k)\\
        -t_x(\theta_k) t_z(\theta_k)&-t_y(\theta_k) t_z(\theta_k) & 1-t^2_z(\theta_k)
      \end{bmatrix} \vec{e}(\theta_k)
    \end{multlined}
\end{align}
And its $q_c$-weighted Euclidean norm,
\begin{align}
\rebuttal{\Vert \vec{e}^c(\theta_k) \Vert_{q_c}^2} &= 
\rebuttal{\vec{e}^c(\theta_k)^T} \bm{Q_c}
  \rebuttal{\vec{e}^c(\theta_k)}
   \label{eq:qc_norm}
\end{align}
where $\bm{Q_c} = q_c \cdot \bm{I}^{3 \times 3}$.

%% file: tikz/tikz_3D_errors_explanation.tex
\begin{tikzpicture}[x={(0.866cm,-0.5cm)}, y={(0.866cm,0.5cm)}, z={(0cm,1cm)}, scale=1.0,
  >=stealth, %
  inner sep=0pt, outer sep=2pt,%
  axis/.style={thick,->},
  wave/.style={thick,color=#1,smooth},
  polaroid/.style={fill=black!60!white, opacity=0.3},
  ]
  \colorlet{darkgreen}{green!40!black}
  \colorlet{lightgreen}{green!80!black}
  \colorlet{darkred}{red!50!yellow}
  \colorlet{lightred}{red!80!black}
  \colorlet{darkblue}{blue!90!black}

  \coordinate (O) at (0, 0, 0);

  \draw[line width=0.4mm, darkblue, dashed] (0.5,0.0,1.0) to[out=90,in=150]
  coordinate[pos=0.6] (A1)
  coordinate[pos=1.7] (B1)
  (0.6,3.0,0) node[right] {};

  \draw[->, line width=0.4mm, darkblue] (0.6,3.0,0) to[out=-30,in=270]
  coordinate[pos=0.0] (A2)
  coordinate[pos=0.35] (X2)
  coordinate[pos=0.8] (B2)
  (5,3,3) node[right] {};
  
  \node[text=darkblue] at ([xshift=18.5]B2) {$\bm{\vec{p}^d(\theta)}$};

  \fill[darkblue] (A2) circle (1.75pt);
  \node[text=darkblue] at ([yshift=9.8, xshift=-4.0]A2) {\footnotesize$\bm{\hat{\theta}_k}$};
  \fill[black, opacity=0.3] (2.7, 2.2, -0.4) circle (3.0pt);
  \node[text=black, opacity=0.9] at (2.7, 2.2, -0.7) {\footnotesize$\bm{\vec{p}_k}$};

  \fill[blue] (X2) circle (1.75pt);
  \node[text=blue] at ([xshift=6.0,yshift=7.8]X2) {\footnotesize$\bm{\theta^*_k}$};

  \draw [line width=0.4mm](0.5,1.5,-1.0) -- (0.5,1.5,1.0) node [above, sloped, midway]{} -- (0.5, 4.5, 1.0) -- (0.5, 4.5, -1.0) -- cycle;
  \node[text=black, rotate=30] at ([xshift=-12.5, yshift=-16.5] 0.5, 4.5, 1.0) {\scriptsize$\bm{\pi(\hat{\theta}_k)}$};

  \draw[line width=0.4mm, darkgreen, <->] (X2) -- coordinate[pos=0.5] (C1) (2.7, 2.2, -0.4) node [right] {};
  \node[text=darkgreen] at ([xshift=10.0]C1) {\footnotesize$\bm{e^c_k}$};

  \draw[line width=0.4mm, darkred, <->] ([yshift=4.0]0.6, 3.0, 0) to[out=-30, in=172] coordinate[pos=0.5] (D1)([yshift=4.0]X2) node[right] {};

  \node[text=darkred] at ([yshift=9.0, xshift=-1.0]D1) {\footnotesize$\bm{e^l_k}$};

  \draw[line width=0.4mm, darkgreen, dashed, ->] (0.6, 3.0, 0) -- (0.6, 2.2,  -0.4) node [label={[xshift = 0.14cm, yshift=0.2cm, rotate=50]$\footnotesize\bm{\vec{\hat{e}^c}(\hat{\theta}_k)}$}] {};
  \draw[line width=0.4mm, darkred, dashed, ->] (0.6, 2.2, -0.4) -- (2.7, 2.2,  -0.4) node [label={[xshift = -1.2cm, yshift=-0.15cm, rotate=-30]$\footnotesize\bm{\vec{\hat{e}^l}(\hat{\theta}_k)}$}] {};
  
  \draw[line width=0.4mm, lightred, dashed, ->] (0.6, 3.0, 0) -- (2.7, 2.2,  -0.4) node [label={[xshift=-0.45cm, yshift = 0.7cm, rotate=-50] $\footnotesize\bm{\vec{e(\hat{\theta}_k)}}$}] {};

\end{tikzpicture}

%% file: sections/application_quads.tex
\section{Application to quadrotors}
To apply our MPCC method to a quadrotor, in this section we define its state and input space, and its dynamics.
\subsection{Quadrotor Dynamics}
\label{sec:quad_dynamics}
The quadrotor's state space is described from the inertial frame $I$ to the body frame $B$, as $\vec{x} = [\vec{p}_{IB}, \vec{q}_{IB}, \vec{v}_{IB}, \vec{w}_{B}]^T$ where $\vec{p}_{IB}\in \mathbb{R}^3$ is the position, $\vec{q}_{IB} \in \mathbb{SO}(3)$ is the unit quaternion that describes the rotation of the platform, $\vec{v}_{IB} \in \mathbb{R}^3$ is the linear velocity vector, and $\vec{\omega}_{B} \in \mathbb{R}^3$ are the bodyrates in the body frame.
The input of the system is given as the collective thrust $\vec{f}_B = \mat{0 & 0 & f_{Bz}}^T$ and body torques $\vec{\tau}_B$.
For readability, we drop the frame indices as they are consistent throughout the description.
The dynamic equations are
\begin{gather}
\begin{aligned}
\dot{\vec{p}} &= \vec{v} & \quad
\dot{\vec{q}} &= \frac{1}{2} \vec{q} \odot \mat{0 \\ \vec{\omega}} \\
\dot{\vec{v}} &= \vec{g} + \frac{1}{m} \mathbf{R}(\vec{q}) \vec{f}_T &
\dot{\vec{\omega}} &= \mathbf{J}^{-1} \left( \vec{\tau} - \vec{\omega} \times \mathbf{J} \vec{\omega} \right)
\label{eq:quad_dynamics}
\end{aligned}
\end{gather}
where $\odot$ represents the Hamilton quaternion multiplication, $\mathbf{R}(\vec{q})$ the quaternion rotation, $m$ the quadrotor's mass, and $\mathbf{J}$ the quadrotor's inertia.

Additionally, the input space given by $\vec{f}$ and $\vec{\tau}$ is decomposed into single rotor thrusts $\vec{f} = [f_1, f_2, f_3, f_4]$ where $f_i$ is the thrust at rotor $i \in \{ 1, 2, 3, 4 \}$.
\begin{align}
\vec{f}_T &= \mat{0 \\ 0 \\ \sum f_i} &
\text{and }
\vec{\tau} &=
\mat{l/\sqrt{2} (f_1 + f_2 - f_3 - f_4) \\
l/\sqrt{2} (- f_1 + f_2 + f_3 - f_4) \\
c_\tau (f_1 - f_2 + f_3 - f_4)}
\label{eq:quad_inputs}
\end{align}
with the quadrotor's arm length $l$ and the rotor's torque constant $c_\tau$.

Furthermore, in order to approximate the most prominent aerodynamic effects, we extend the quadrotor's dynamics to include a linear drag model \cite{Faessler18ral}. Let $\bm{D}$ be the diagonal matrix with drag coefficients $d$ such that $\bm{D} = \text{diag}\left(d_x, d_y, d_z\right)$, we expand the dynamic model from \eqref{eq:quad_dynamics} by:
\begin{equation}
\dot{\bm{v}} = \bm{g} + \frac{1}{m} \bm{R}(\bm{q}) \bm{f_T} - \bm{R}(\bm{q}) \bm{D} \bm{R}^\intercal(\bm{q}) \cdot \bm{v}
\end{equation}

\subsection{Optimal Control Problem Formulation}
In this section, we propose to include the dynamics introduced in section \ref{sec:quad_dynamics} within the MPCC formulation outlined in problem \eqref{eq:ocp_generic}. To do this, the state space described in \eqref{eq:quad_dynamics} needs to be augmented to include progress dynamics. 
A direct way to achieve this is to add $\theta$ as a state and $v_{\theta}$ as a virtual input. However, this way there would be no limit in how fast $v_{\theta}$ changes and infinite changes would be allowed leading to very noisy control inputs. To alleviate this, and following typical implementations of contouring controllers \cite{Liniger_Domahidi_Morari_2015, Schwarting_Alonso-Mora_Paull_Karaman_Rus_2018}, we use the \emph{progress acceleration}, $\Delta v_{\theta} = \frac{d v_{\theta}}{ d t}$ as a virtual input instead. The resulting augmented state and input spaces are shown in \eqref{eq:state_space}.

\begin{align}
  \vec{x} = \begin{bmatrix}\vec{p}& \vec{q}& \vec{v}& \vec{w} & \vec{f} & \theta & v_{\theta} \end{bmatrix}^T, \qquad \vec{u} = \begin{bmatrix}\Delta v_{\theta} & \Delta\vec{f}\end{bmatrix}^T
  \label{eq:state_space}
\end{align}

where the dynamics of the augmented states are:

\begin{align}
  \vec{f}_{k+1} &= \vec{f}_k + \Delta \vec{f}_{k} \Delta t \notag \\
  \theta_{k+1} &= \theta_k + v_{\theta, k} \Delta t \notag \\
  v_{\theta, k+1} &= v_{\theta, k} + \Delta v_{\theta, k} \Delta t
  \label{eq:augmented_dynamics}
\end{align}

Thus, the full optimal control problem is as follows:

\begin{align}
  \pi(\vec{x}) = & \argmin_u
  & & \begin{multlined}\sum_{k = 0}^{N } \Vert \rebuttal{\vec{e}^l(\theta_k)} \Vert_{q_l}^2 + \Vert \rebuttal{\vec{e}^c(\theta_k)} \Vert_{q_c}^2 + \Vert\vec{\omega}_k\Vert_{\bm{Q_{\omega}}}^2 \\ + \Vert\Delta v_{{\theta}_k}\Vert_{r_{\Delta v}}^2 + \Vert\Delta\vec{f}_k\Vert_{\bm{R_{\Delta f}}}^2 - \mu v_{\theta,k}
  \end{multlined} \notag\\
  & \text{subject to} & & \vec{x}_0 = \vec{x} \notag \\
  &&&\vec{x}_{k+1} = f(x_k, u_k) \notag \\
  &&&\lb{\vec{\omega}} \leq \vec{\omega} \leq \ub{\vec{\omega}} \notag \\
  &&&\lb{\vec{f}} \leq \vec{f} \leq \ub{\vec{f}} \notag \\
  &&& 0 \leq v_{\theta} \leq \ub{v_{\theta}} \notag \\
  &&& \lb{\Delta v}_{\theta} \leq \Delta v_{\theta} \leq \ub{\Delta v_{\theta}} \notag \\
  &&& \lb{\Delta \vec{f}} \leq \Delta \vec{f} \leq \ub{\Delta \vec{f}}
  \label{eq:full_ocp}
\end{align}

where the lag and countour norms were defined in section \ref{sec:lag_contour_errors}, in \eqref{eq:ql_norm} and \eqref{eq:qc_norm}, respectively, and the dynamics $f(x_k, u_k)$ come from the discretization of \eqref{eq:quad_dynamics} and \eqref{eq:quad_inputs}, and from \eqref{eq:augmented_dynamics}.

In problem \eqref{eq:full_ocp}, constraints and costs on $\Delta v_\theta$ and $\Delta \vec{f}$ have been added such that the solver does not have complete freedom of choosing them arbitrarily. 
These extra constraints and costs are needed in contouring control formulations to provide stability to the optimization problem \cite{Schwarting_Alonso-Mora_Paull_Karaman_Rus_2018, Liniger_Domahidi_Morari_2015}. 
Without them, the outputs of the solver are very noisy, rendering the inputs unusable for real-world applications. The cost on $\vec{\omega}$ has also proven to be necessary for the stability of the solution since it forces the solver to keep low body rates whenever possible. 
Furthermore, this cost also allows tracking a reference in $\omega_z$ to control the platform's orientation externally. Lastly, the limits on $v_{\theta}$ ensure non-reversal of the reference path and set a maximum speed at which the path is traversed. The complete MPCC pipeline is described in Algorithm \ref{alg:MPCC}.

\SetKwFor{Loop}{loop}{}{end}
\begin{algorithm}
  \SetAlgoLined
  \KwIn{$p^d(t)$ the 3D path to track, $x_0$ initial state}

  $\mathcal{P} \gets [\enskip]$ vector of corresponding positions, velocities and arc lengths\\
  \BlankLine
  \BlankLine
  \uIf{$p^d(t)$ is a sequence of sampled points} {
    $\mathcal{P}(i) \gets$ assume linearity to compute arc length}
  \uElseIf{$p^d(t)$ is continuous} {
    $\mathcal{P}(i) \gets$ use \emph{bisection method} from \cite{ArcLengthWang2002}
  }
  \BlankLine
  \BlankLine
  \For{$i \in [1, P]$}{
    $\rho_i(\theta_k) \gets \mathcal{P}(i), \mathcal{P}(i - 1)$ compute splines from points \\
  }
  $p^d(\theta_k) \gets \{\rho_i(\theta_k)\}_{i = 0}^P$ as in \eqref{eq:arc_length}
  \BlankLine
  \BlankLine
  $\{x_k\}_{k=0}^N \gets x_0$, $\{\theta_{k}\}_{k = 0}^N \gets 0$ \\
  \Loop{}{
    $\mathcal{C}(\theta_k) \gets [\enskip]$ vector of spline coefficients\\
    \For{$k \in [0, N]$} {
      $\theta_{pred,k} \gets$ $\theta_k$ from previous $X^*_{pred, k}$\\
      $\mathcal{C}(\theta_k) \gets p^d(\theta_{pred,k})$ \\
    }
    compute \eqref{eq:ql_norm} and \eqref{eq:qc_norm} given $\mathcal{C}(\theta_k)$ \\
    $U^*_{pred}, X^*_{pred} \gets $ solve problem \eqref{eq:full_ocp} \\
    apply $U ^ *_{pred, 0}$ to the platform \\
  }

  \caption{MPCC}
  \label{alg:MPCC}

\end{algorithm}

%% file: sections/dynamic_allocation.tex
\subsection{Dynamic Allocation of Contouring Weight}
\label{sec:dyn_weight}
Given a set of $M$ waypoints located at $\{\vec{p}_{g, j}(\theta_k)\}_{j=0}^M$, we consider the problem of passing through them in the shortest time possible. When relating this to the optimization problem described in \eqref{eq:full_ocp}, one can notice two competing objectives in the controller cost function: the contouring weight $q_c$, which pushes the platform towards the center of the 3D path and eventually through the \changeElia{waypoints}, and the progress weight $\mu$, which makes it traverse the path as fast as possible.

Because of these competing terms in the cost function, there is a need to find a balance between going fast and reducing the contour error. For waypoint tracking, however, we are only concerned about the tracking performance to the 3D path in the points where the waypoints are. For this, the position of the waypoints $\{\vec{p}_g\}_{j=0}^M$ need to be included in our optimal control formulation \eqref{eq:full_ocp}. This can be done in different ways.

On the one hand, the position constraints could be directly included in the optimization problem, i.e., adding
\begin{align*}
    \Vert\vec{p_k} - \vec{p}_{g}(\theta_j) \Vert_2^2 \leq d_{tol,j} \quad \forall j \in [0, M]
\end{align*}
to problem \eqref{eq:full_ocp}. However, this approach has been found to be unsuccessful. 
If the constraints are added as hard constraints, the problem quickly becomes unfeasible. 
If they are soft constraints, the controller behavior becomes extremely conservative. 
The optimization problem becomes very complex because we need to add a slack variable, a constraint for the slack variable, a cost for the slack variable, and the soft constraint itself. 

On the other hand, the contour weight $q_c$ could be dynamically allocated online. Instead of keeping it constant, we make it depend on the desired position of the platform $q_c(\theta_k) = q_c(\vec{p}^d(\theta_k))$ such that when we are close to a waypoint, the contour error is the largest one in the cost function, and everywhere else we give more importance to the progress. The mapping that has been chosen is a collection of 3D Gaussians placed at the waypoint positions $\vec{p}_{g,j}(\theta_k)$, yielding
\begin{equation*}
    q_c(\vec{p}^d(\theta_k)) =
    \sum_{j = 0}^M \frac{e^{\left(-\frac{1}{2}\left(\vec{p}^d(\theta_k) - \vec{p}_{g,j}(\theta_k)\right)^T \Sigma^{-1} \left(\vec{p}^d(\theta_k) - \vec{p}_{g,j}(\theta_k)\right)\right)}}{\sqrt{(2\pi)^3 \vert\Sigma\vert}}
\end{equation*}
where $\Sigma$ is the diagonal covariance matrix that indicates how wide the Gaussians are in $x$, $y$, and $z$ axes. The elements of $\Sigma$ are chosen to be equal in the three axes and relatively small such that (i) they only contain information around the waypoint position and (ii)  the Gaussians do not overlap with each other.

Using Gaussians as the waypoint weighting function has been traditionally used in the literature for similar purposes, such as \cite{waypoint_gaussian_siegwart}. For this work, the main reasons behind the decision of using Gaussians are:
\begin{itemize}
    \item As they are smooth curves, arbitrarily small changes in $\theta_k$ lead to arbitrarily small changes in $q_c(\vec{p}^d(\theta_k))$, preventing discontinuities in the cost function and therefore helping with controller stability.
    \item As explained in section \ref{sec:problem_formulation}, there is always a lag error. Therefore the projection from our current position $p_k$ to the desired path $\vec{p}^d(\theta_k)$ might not be completely accurate at all times, and this approximate projection might both be lagging or leading. The symmetry properties of Gaussians ensure that the costs are attained equally no matter in which direction these projection inaccuracies lie.
\end{itemize}

\begin{figure}[tp]
\centering
\input{tikz/tikz_gaussian}
\includegraphics[width=0.7\linewidth]{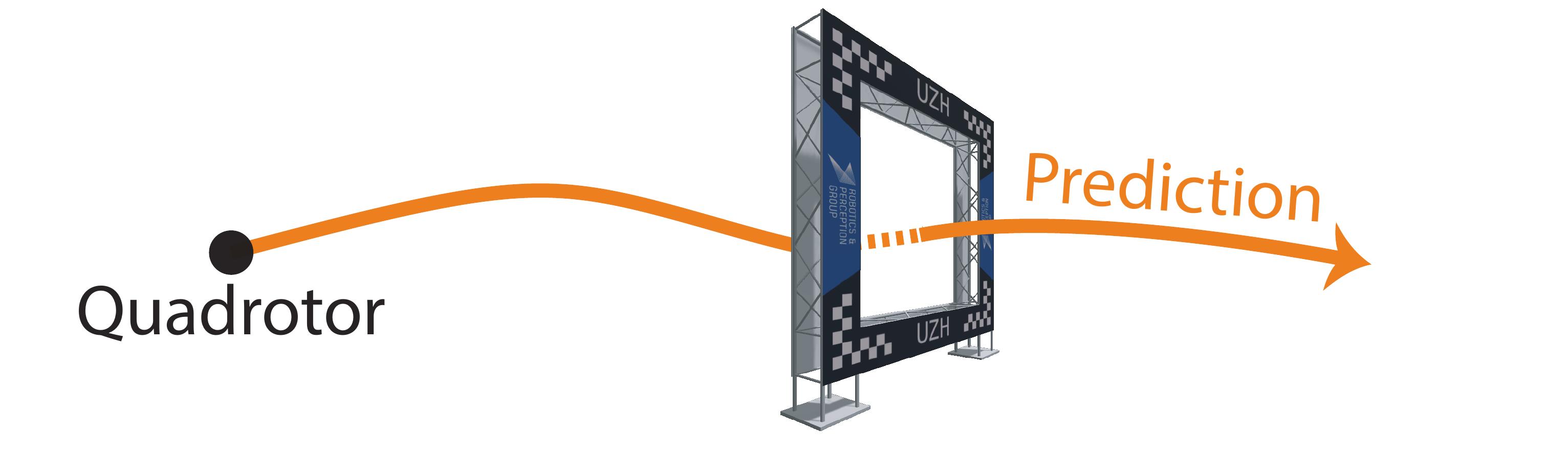}
\caption{\changeSix{Top: }dynamic allocation of the weight in the $x$ axis where the reference path $p_x^d(\theta_k)$ (shown in black) is the $x$ component of $\vec{p}^d(\theta_k)$. When closer to the waypoint, the contouring weight $q_c(p_x^d(\theta_k))$ (shown in blue) grows to $q_{wp}$, decreasing the tracking error around the waypoints. Everywhere else $q_c$ is kept constant and equal to $q_{nom}$ such that the MPCC has more freedom to optimize for progress. \changeSix{Bottom: Illustration of how the contouring weight allocation would translate to for the drone racing task, where the gate position corresponds to the position of the waypoint.}
}
\label{fig:gaussians}
\end{figure}

A 1D illustrative example in Fig. \ref{fig:gaussians} shows how this looks like for the $x$ axis. Note that the contouring cost is a function of the desired path $\vec{p}^d(\theta_k)$ and not of the current position $p_k$. This way, we ensure that these costs will always be attained since the reference path will always pass through the waypoints. \rebuttal{A comparison study of how the platform behaves with and without this dynamic allocation of $q_c$ in a simulated environment is shown in Fig. \ref{fig:min_snap_no_gates}.}

Summarizing, the proposed controller takes as inputs a 3D path $\vec{p}^d(\theta)$ and a set of waypoints $\left\{\vec{p}_g\right\}_{j = 0}^M$ and outputs a policy that optimally balances going as fast as possible and being close to the tracked path where the waypoints are. It is therefore interesting to see how its performance compares with a true time-optimal trajectory, even if we expect to encounter differences due to (i) the horizon of the MPCC is limited, (ii) several approximations have been made in the proposed formulation, and (iii) there are additional terms in the cost function of problem \eqref{eq:full_ocp} apart from purely maximizing the progress.

%% file: tikz/tikz_gaussian.tex
\newcommand\gauss[2]{1/(#2*sqrt(2*pi))*exp(-((x-#1)^2)/(2*#2^2))} 
\begin{tikzpicture}
\begin{axis}[axis x line=none, axis y line=none, axis x line shift=-0.06cm, axis line style = thick, xtick=\empty, ytick=\empty, every axis plot post/.append style={
    mark=none,domain=-2.5:2.5,samples=50,smooth, thick},
  xmin=-3,xmax=3,ymin=0,ymax=1.5,
  xlabel={}, ylabel={}, xlabel style={xshift=0.45cm, yshift=0.5cm}] 
\draw[->, very thick] (axis cs: -2.0, 0.7) to [out = 0, in=180]
    coordinate[pos=0.0] (x0)
    coordinate[pos=0.2] (x1)
    coordinate[pos=0.4] (x2)
    coordinate[pos=0.6] (x3)
    coordinate[pos=0.8] (x4)
    coordinate[pos=1.0] (xend)
    (axis cs: 2.0, 0.3);
\fill[black] (x0) circle (2pt);
\fill[black] (x1) circle (2pt);
\fill[black] (x2) circle (2pt);
\fill[black] (x3) circle (2pt);
\fill[black] (x4) circle (2pt);

\node[text=black] at ([yshift=0.25cm]x0) {\tiny $\boldsymbol{k{=}0}$};
\node[text=black] at ([yshift=0.25cm]x1) {\tiny $\boldsymbol{k{=}1}$};
\node[text=black] at ([yshift=0.25cm]x2) {\tiny $\boldsymbol{k{=}2}$};

\node[text=black] at ([yshift=0.4cm]xend) {$\boldsymbol{p^d_x(\theta_k)}$};
\addplot [blue, opacity=0.5] {\gauss{0.35}{0.35} + 0.15};
\node[text=blue] at (axis cs: 1.65, 1.1) {$\boldsymbol{q_c(p^d_x(\theta_k))}$};
\draw [dashed, opacity=0.4] (x0) to ([yshift=-2.09cm]x0);
\draw [dashed, opacity=0.4] ([yshift=3.2cm]x3) to (x3);
\node[text=black] at ([yshift=-0.4cm, xshift=0cm]x3) {\tiny $\boldsymbol{p^d_{x,g}(\theta_j)}$};

\fill[blue] ([yshift=-2.09cm]x0) circle (2pt);
\node[text=blue] at ([yshift=-1.9cm, xshift=-0.5cm]x0) {\footnotesize $\boldsymbol{q_{nom}}$};

\draw [dashed, opacity=0.4] (x1) to ([yshift=-1.92cm]x1);
\fill[blue] ([yshift=-1.92cm]x1) circle (2pt);

\draw [dashed, opacity=0.4] (x2) to ([yshift=-1.0cm]x2);
\fill[blue] ([yshift=-1.0cm]x2) circle (2pt);

\fill[blue] ([yshift=3.18cm]x3) circle (2pt);
\node[text=blue] at ([yshift=3.25cm, xshift=-0.5cm]x3) {\footnotesize $\boldsymbol{q_{wp}}$};

\draw [dashed, opacity=0.4] (x4) to ([yshift=-0.3cm]x4);
\fill[blue] ([yshift=-0.3cm]x4) circle (2pt);

\end{axis}
\end{tikzpicture}

%% file: sections/path_generation.tex
\rebuttal{
\section{Path Generation}
\label{sec:path_generation}
In this section, we introduce three different ways of generating the nominal path that is used by the proposed MPCC, $\vec{p}^d(t)$. Even though the proposed paths in this section are parameterized by time, this is not necessary. The only requirement for the nominal path is that it is continuously differentiable with respect to its parameter. As described in Section \ref{sec:arc_length_splines}, for their use with the proposed MPCC algorithm, these paths first need to be reparameterized by arc-length to obtain $\vec{p}^d(\theta_k)$.
\subsection{Multi-Waypoint Minimum Snap}
\label{sec:pg_min_snap}
Polynomial, and in particular minimum snap trajectories, have been used for full state trajectory planning for \changeFifth{quadrotors} in different applications \cite{Mellinger11icra}, \cite{Mueller13iros}. 
One drawback of polynomial trajectories is that states and inputs are sampled from their derivatives and, therefore, can only offer polynomial control inputs. 
Since polynomials are smooth and can only attain their maximum value at single points, it is not possible to get arbitrarily fast-changing control inputs. 
This undermines the agility and aggressiveness that can be achieved by the platform, rendering them suboptimal for minimum-time applications. 
In this section, we use polynomial trajectory planning to obtain a continuously differentiable 3D path $\vec{p}^d(t)$ that passes through the gates. The time allocation of the control inputs and of the states of the drone will be generated by the \change{MPCC} controller at \changeElia{runtime.}

These polynomial trajectories have been generated in a receding horizon fashion. 
Starting from the current position of the platform, we generate a polynomial that \emph{(i)} passes through the $N$ next waypoints and \emph{(ii)} minimizes the second derivative of the acceleration (\change{or snap, as in \cite{Mellinger11icra}}). 
From this result, we select only the part from the current waypoint to the next waypoint and re-run the planning for the next waypoint with the same horizon length. 
This approach does not require an initial selection of speeds at each waypoint, as it is required for polynomial trajectories.
Additionally, it keeps the order of the polynomial segments lower as compared to generating one single polynomial trajectory that passes through all the gates and hence avoids numerical issues. 
Notice that there is no notion of time minimization in the way we generate this minimum snap trajectory, meaning that the time-parameterized reference can be relatively slow. 
However, this is not relevant when tracking it with MPCC because the reference is first reparameterized by arc length.
From this point on, this polynomial trajectory will be referred to as \emph{minimum snap trajectory}.
\subsection{Time-Optimal Full Model}
\label{sec:pg_cpc}
This path generation algorithm generates a time-optimal trajectory that passes through a sequence of given waypoints \cite{foehn2021CPC}. 
This is achieved by introducing a formulation based on \emph{Complementary Progress Constraints (CPC)} along the trajectory, which allows simultaneous optimization of the time allocation and of the trajectory itself.
Because it takes into account the full non-linear model of the quadrotor platform down to single rotor thrusts, the generated trajectory is able to exploit the full actuator potential at all times.
This algorithm sets the state-of-the-art on time-optimal planning for quadrotors.
However, different sources of approximation should also be noticed. First, inside the optimization problem, the trajectory itself is discretized and divided into a finite number of nodes. Second, even if the model of the platform is very accurate, it is not perfect, and model mismatches are definitely present.
Nevertheless, the resulting trajectories push the platform to the limits of what is achievable for the given model, and are even able to beat professional drone pilots in the challenging task of drone racing \cite{foehn2021CPC}.

The main drawback of using a time-optimal CPC trajectory as the reference is that they take a significant amount of time to compute.
The generation time of these full state time-optimal trajectories ranges from minutes to several hours, making them intractable for real-time applications.
One attempt to reduce this computation time has been made \cite{Song_Steinweg_Kaufmann_Scaramuzza_2021} using reinforcement learning (RL), but, although this approach scales better with track length, training times are still of the order of hours.
\change{Instead}, the proposed MPCC method does not need most of the information included in these time-optimal trajectories, and it suffices to provide a 3D path $\vec{p}^d(t)$ that is continuously differentiable.
Fig. \ref{fig:solve_times_CPC_RL_PMM} depicts how the computation times to generate these trajectories using different approaches evolve with the track length.
\begin{figure}[tp]
  \centering
  \includegraphics[width=\linewidth]{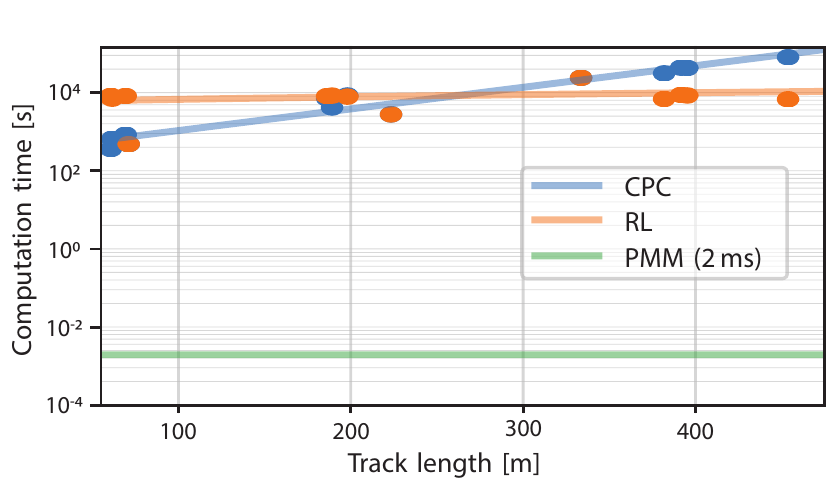}
  \caption{Evolution of computation times with respect to track length for optimal time trajectory generation. Both available approaches so far, model-based (CPC) and learning-based (RL), are not usable in \changeElia{applications that require online re-planning}, whereas the proposed contouring controller allows us to generate optimal trajectories using a point-mass model (PMM), rendering computation times suitable for real-time re-planning (\SI{2}{\milli\second}).}
  \label{fig:solve_times_CPC_RL_PMM}
\end{figure}
\subsection{Time-Optimal Point-Mass Model}
\label{sec:pg_pmm}
As mentioned, one of the benefits of MPCC is that it can track non-feasible paths.
Throughout this paper we exploit this property by generating 3D paths that do not consider the quadrotor dynamics, but instead a simpler point-mass model (PMM) where the drone is assumed to be a point mass with bounded accelerations as inputs.
This significantly reduces the computational burden since the generation of point-to-point trajectories using a point mass model has a closed-form solution.

To generate these PMM time-optimal paths, the same sampling approach that was used in \cite{Foehn20rss} is employed here.
Given an initial state consisting of position and velocity, and given the reduced dynamics of a point-mass model, $\ddot{\vec{p}} = \vec{u}$, with the input acceleration being constrained $\lb{\vec{u}} \leq \vec{u} \leq \ub{\vec{u}}$, it can be shown by using Pontryagin's maximum principle \cite{bertsekasDP} that the time-optimal control input results in a bang-bang policy. 
Furthermore, if we also add constraints on the velocity, $\lb{\vec{v}} \leq \vec{v} \leq \ub{\vec{v}}$, the time optimal control policy has a bang-singular-bang solution \cite{Maurer} of the form:
\begin{align}
  u_x^*(t) = \left\{
  \begin{matrix}
    \lb{u}_{x}, & 0 \leq t \leq t_1^* \\
    0,  & t_1^* \leq t \leq t_2^* \\
    \ub{u}_{x},  & t_2^* \leq t \leq T_x^*
  \end{matrix}\right.
  \label{eq:time_optimal_x}
\end{align}
or vice versa, starting with $\ub{u}_{x}$. 
It is straightforward to verify that there exists a closed-form solution for the switching times. 
In \eqref{eq:time_optimal_x} we depict the solution for the $x$ axis, but this can be extended for $y$ and $z$ axes without loss of generality.

Once we have the per-axis minimum time policy, we choose the maximum of these minimum times \change{\mbox{($T^* = \text{max}(T^*_x, T^*_y, T^*_z)$)}} and slow down the other two axes' policies to make all three axes' \change{durations equal.}
\change{To do this, a new parameter $\alpha \in [0, 1]$ is introduced that scales the acceleration bounds. For example, if we need to increase the duration of the $x$ axis, the applied control inputs are scaled to $\alpha \lb{u}_x$ and $\alpha \ub{u}_x$, respectively.}
For more details on how these trajectories are sampled, the reader is referred to \cite{Foehn20rss}.

By using the described method, it is possible to generate a time-optimal point-to-point motion primitive for every fixed start and end positions and velocities.
In order to find the time-optimal, multi-waypoint trajectory efficiently, the path planning problem is interpreted as a shortest path problem.
This reduces the path planning problem to finding the drone’s optimal state at each gate such that the total time is minimized.
At each gate, $M$ different velocities are sampled at random such that they lie in a cone pointing towards the exit direction of the gate.
The cost from each sampled state at the previous gate is set to be equal to the duration, $T^*$, of the time-optimal motion primitive that guides the drone from one state to the other.
Due to the existence of a closed-form expression for the minimum time, $T^*$, setting up and solving the shortest path problem can be done very efficiently using, e.g., Dijkstra’s algorithm \cite{bertsekasDP}, resulting in the optimal path $\vec{p}^d(t)$ that we can then track with the proposed MPCC.
In order to further reduce the computational cost, the path is planned in a receding horizon fashion, i.e., the path is only planned through the next $H_g$ waypoints.
Fig. \ref{fig:sampling} shows the resulting trajectory when applying this method to a sequence of 7 waypoints.
It is worth noting that, because the PMM trajectory is not feasible, a reference tracking controller such as MPC fails to track it.
}

%% file: sections/hover_to_hover.tex
\section{Simulation Experiments}
In this section, we design a series of simulation experiments aimed to demonstrate the capabilities of our method.
First, in section \ref{sec:hover_hover} we show a comparison \change{between} our method \change{and} a true time-optimal trajectory for the simple task of hover to hover flight. 
Then, in section \ref{sec:path_generation_drone} we evaluate the performance of our method in the more challenging task of \change{multi-waypoint flight applied to drone racing.}
In this context, we design different ways of generating the 3D path $\vec{p}^d(\theta)$ that passes through all the waypoints and compare them in terms of speed and lap times with respect to the main baseline, \cite{foehn2021CPC}. 
\changeThird{Finally, in section \ref{sec:delay_ablation} we evaluate the robustness of the proposed MPCC controller against the presence of time delay in the loop and compare its performance to that of a standard MPC controller.}
Note that, while in this section we focus only on simulation experiments, in Section \ref{sec:exp_realworld} we present the real-world results.

\change{The quadrotor configuration used throughout this paper, denoted \emph{RPG Quad} in Table \ref{tab:quads}, represents the model parameters of our in-house designed drone.}

\newcolumntype{C}[1]{>{\centering\arraybackslash}m{#1}}
\begin{table}[tb]
\caption{\change{Quadrotor Parameters}}
\label{tab:quads}
\begin{center}
    \begin{small}
    \setlength{\tabcolsep}{5pt}
    \setlength\extrarowheight{1.5pt}
        \begin{tabular}{C{2.8cm}||c}
            \toprule
            \rowcolor{gray!30!white}
            Property & RPG Quad  \\
            \hline
            $m$ $[\si{\kilo\gram}]$ & $0.85$ \\
            \hline
            $l$ $[\si{\meter}]$ & $0.15$ \\
            \hline
            $\text{diag}(J)$ $[\si{\gram\meter^2}]$ & $[2.5, 2.1, 4.3]$ \\
            \hline
            $[T_{min}, T_{max}]$ $[\si{\newton}]$ & $[0.0, 7.0]$ \\
            \hline
             $c_\tau$ $[1]$ & $0.022$\\
            \hline
            $\omega_{max}$ $[\si{\radian\per\second}]$ & $10$\\
            \bottomrule
        \end{tabular}
    \end{small}
\end{center}
\end{table}


\subsection{Time-Optimal Hover to Hover Comparison}
\label{sec:hover_hover}
\change{To} compare up to which extent the maximization of the progress in problem \eqref{eq:full_ocp} is equivalent to the minimization of the trajectory time, we have set up a simple simulation experiment: to execute a \SI{15}{\meter} hover to hover trajectory in the $x$ axis. 
This task has been used as a benchmark in previous work \cite{foehn2021CPC, Hehn12ar, Loock13ecc}. 
First, we compute the time-optimal trajectory using the same quadrotor model described in section \ref{sec:quad_dynamics}. 
We compute it using the CPC approach \cite{foehn2021CPC}, which takes between 20 and 30 minutes to converge on a desktop computer. 
This trajectory is the theoretical lower bound on how fast this task can be performed. On the other hand, we simulate the platform controlled by the proposed MPCC to follow a 3D straight line reference that can be run in real-time. 
It is important to clarify that no information is shared between the full model time-optimal trajectory and the MPCC controller: the 3D reference path $\vec{p}^d(\theta)$ is plainly an arc-length parametrized straight line. 
In both cases, the maximum acceleration of the platform has been limited to \SI[per-mode = symbol]{20}{\meter\per\second\squared}, \change{as in \cite{foehn2021CPC, Hehn12ar, Loock13ecc}}.

\begin{figure}[tb]
  \centering
  \includegraphics[width=\linewidth]{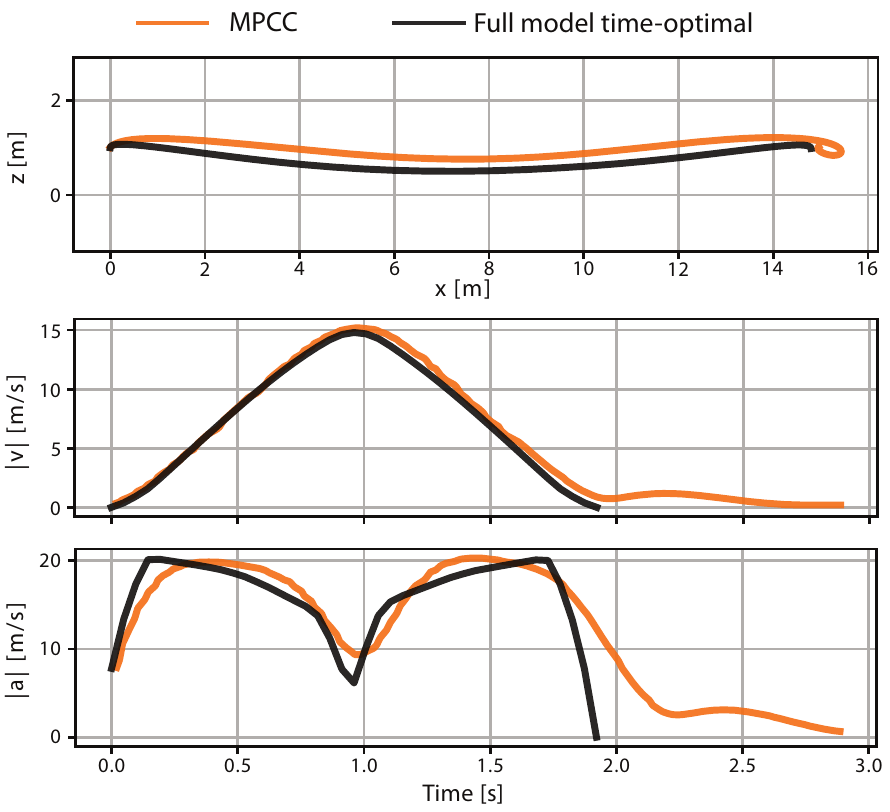}
  \caption{\change{Simulated position, velocity, and acceleration profiles of the MPCC compared to the time-optimal trajectory when executing a 15 meters hover to hover task. Note that the MPCC is solely tracking a straight line between the two hover states. The similarities in the speed and acceleration profiles show how the proposed MPCC controller attempts to find an approximation to the pre-computed time-optimal trajectory.}}
  \label{fig:hover_to_hover}
\end{figure}

In Fig. \ref{fig:hover_to_hover} we show a comparison of position, velocity, and acceleration between the CPC time-optimal reference and the simulated MPCC sequence of states. 
The results of this simulation show that the evolution of the MPCC controlled system is very close to the time-optimal CPC reference. 
The main differences between both approaches can be attributed to various reasons:
\begin{enumerate}[label=(\roman*)]

\item \rebuttal{Due to (a) its limited time horizon and (b), that the cost function \eqref{eq:full_ocp} does not only include maximization of progress, but also minimization of contour and lag errors, general smoothness of the inputs, etc.,} the proposed MPCC method finds locally optimal solutions. In contrast, the CPC algorithm has access to information about the evolution of all the past and future states of the platform and therefore finds the time-optimal trajectory.
\item \change{While the MPCC already shows the simulated results of an executed trajectory, the CPC is a planned reference trajectory. We have chosen to compare directly with the reference because it does not depend on the tuning parameters of an external controller.}
\end{enumerate}

%% file: sections/planning_3D_path.tex
\subsection{Ablation Study on the Choice of Reference Trajectory}
\label{sec:path_generation_drone}
In this section, we perform an ablation study that aims to show and evaluate up to which extent a better choice of the reference path helps the MPCC finding a good approximation of the time-optimal policy. 
\rebuttal{To this end, we use the path generation algorithms introduced in Section \ref{sec:path_generation} and generate paths that pass through a given sequence of waypoints. 
Then, we evaluate how our MPCC algorithm performs in terms of speed and lap time when tracking these different paths.
}

\rebuttal{The locations of the waypoints have been chosen to be the same as in \cite{foehn2021CPC} and their spatial distribution is shown in Fig. \ref{fig:sampling}.}
This track has been designed by professional drone pilots and will be used as the main benchmark throughout this paper, under the name \emph{CPC-track}. 
It is important to note that the same controller tuning and contouring weights have been used for all our simulation experiments.

\subsubsection{Multi-Waypoint Minimum Snap}

\begin{figure}[t]
\centering
\includegraphics[width=\linewidth]{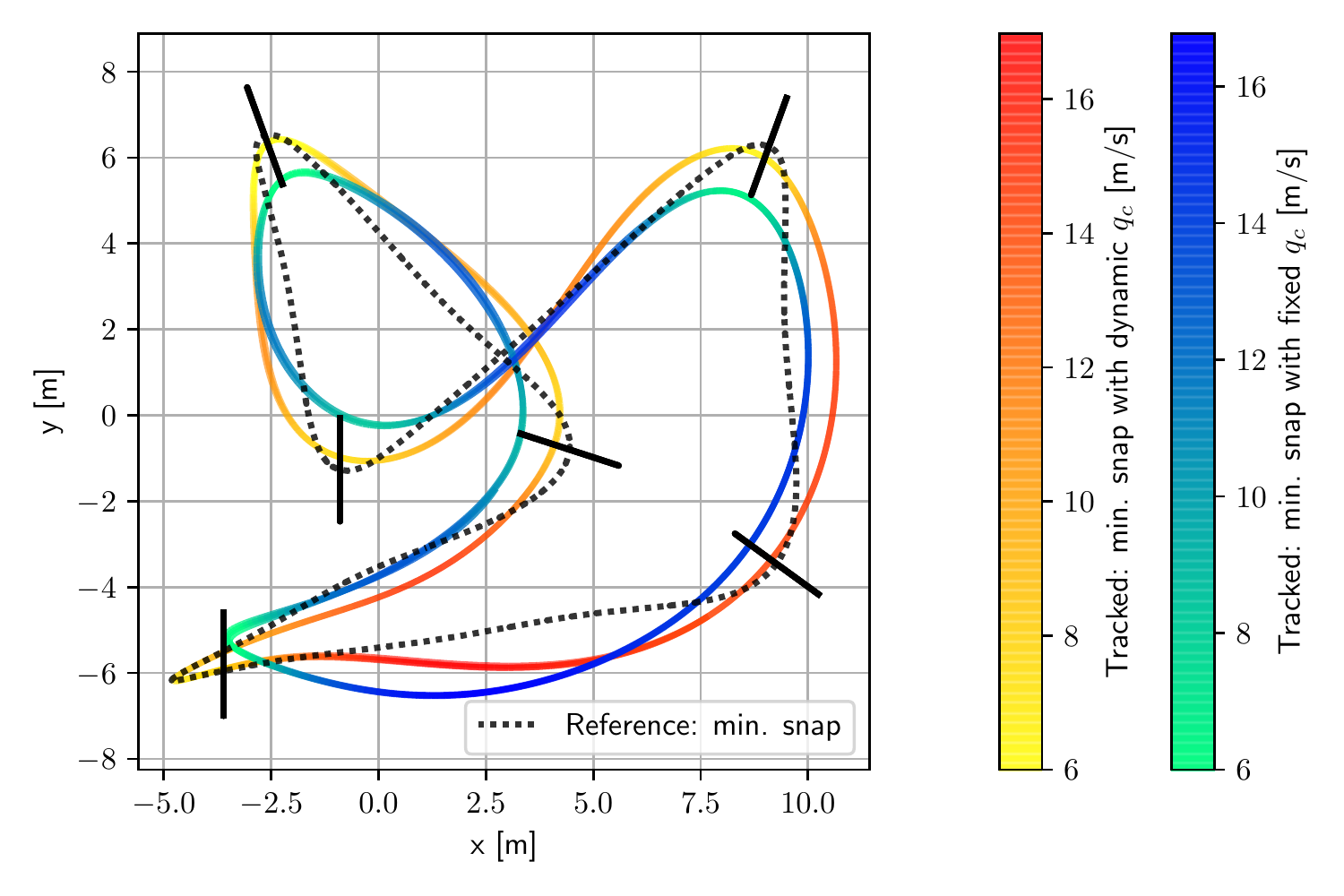}
\caption{\change{\emph{CPC-track} consisting of 7 gates (black segments). A 3D view of this track can be found in Fig. \ref{fig:sampling}. The polynomial reference is shown in dotted black and is tracked using the proposed MPCC, with dynamic and fixed contouring weights $q_c$ respectively. Both tracked trajectories are colored by speed profile (dynamic contouring weight: orange-yellow color, fixed contouring weight: blue-green color). Notice that the tracked trajectory when $q_c$ is fixed does not successfully pass through all the gates. Instead, when $q_c$ changes dynamically, the platform successfully goes through all the gates without any notable change in speed.}}
\label{fig:min_snap_no_gates}
\end{figure}

In Fig. \ref{fig:min_snap_no_gates}, we show an \change{XY view of the MPCC tracking when following a minimum snap trajectory.
To show the benefit of using the dynamic contouring weight method introduced in Section \ref{sec:dyn_weight}, we compare the trajectory execution of the proposed MPCC when $q_c$ is dynamically allocated and when $q_c$ is fixed to a constant value.
If $q_c$ remains fixed, one can notice how the platform does not pass through most of the gates or passes too close to the edge (which would result in a collision with the gate frame in a real flight). 
\changeElia{Since we use the same value for $q_c$ in the entire track, the approach lacks information about the location of the waypoints and treats the entire path equally. 
The balance between maximizing progress and minimizing contour error is constant.}
However, when using the dynamic contouring weight formulation, the platform successfully passes through all the gates. 
Furthermore, \changeElia{even if with the dynamic formulation we deprioritize the maximization of the progress at the location of the waypoints,} it does not have any notable impact in the speed profile, as seen in the colored bars of Fig. \ref{fig:min_snap_no_gates}.}
\change{Therefore, in the remainder} of the paper, a dynamic contouring weight formulation is used.

\subsubsection{\change{Time-Optimal Full Model (CPC)}}
\begin{figure}[htp]
\centering
\includegraphics[width=\linewidth]{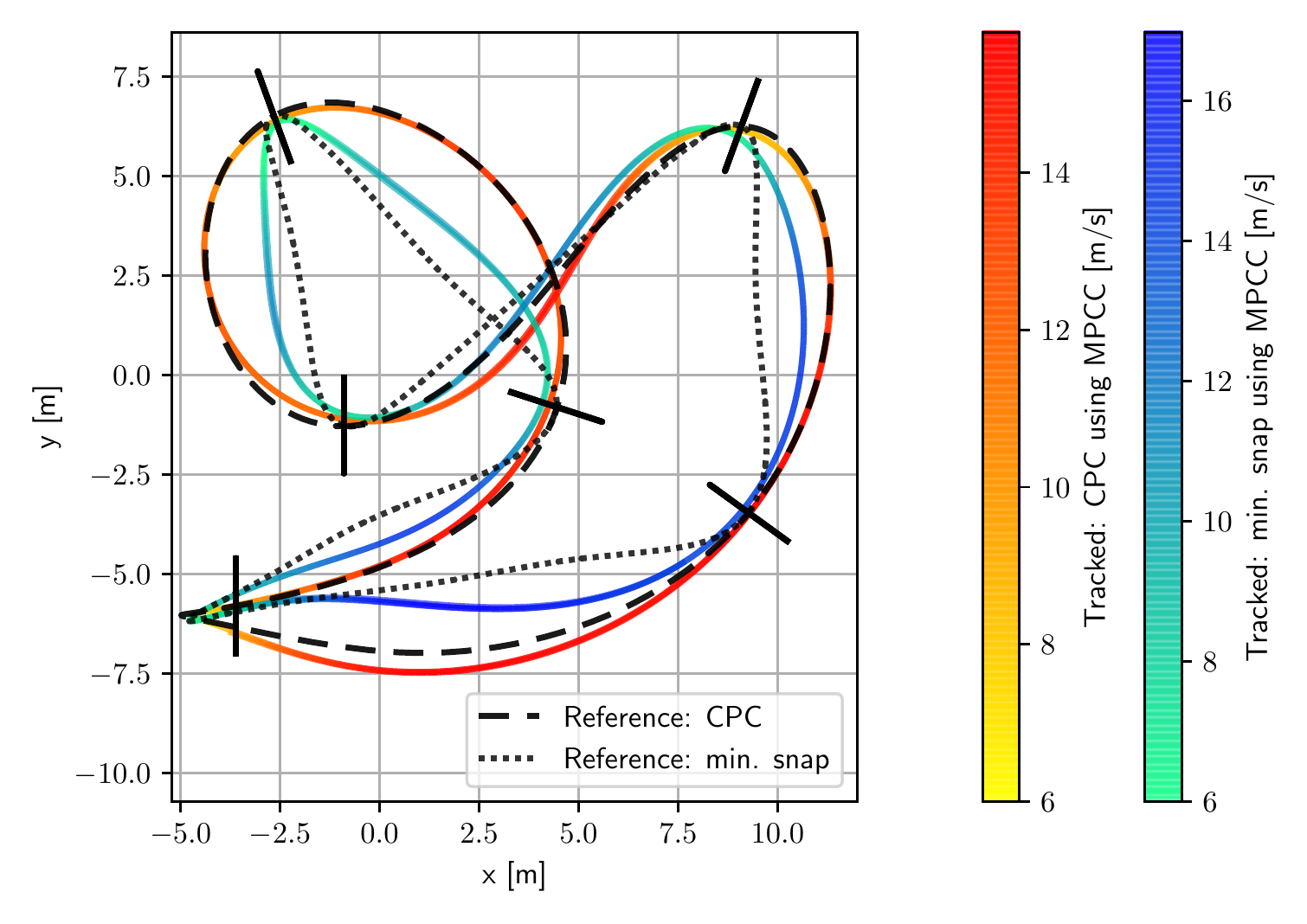}
\caption{Simulated tracking performance comparison on the CPC track when using minimum snap and CPC as a reference. The minimum snap reference is shown in dotted black, and the CPC reference is shown in dashed black, while the corresponding MPCC tracked trajectories are colored by their speed profile (low contouring weight: blue-green color, high contouring weight: orange-yellow color). The MPCC controller is using the dynamic \change{contour} weights shown in Fig. \ref{fig:gaussians}.}
\label{fig:CPC_min_snap_gates}
\end{figure}


In Fig. \ref{fig:CPC_min_snap_gates} we show the executed 3D path and the velocity profile for both the minimum snap reference and the CPC reference. 
\changeElia{Because the minimum snap trajectory \rebuttal{does not minimize} the total trajectory time, it has no notion of the long-term global behavior of the platform, leading to sharper turns and long straight lines.
Because of its finite time horizon, when tracking this minimum snap reference, the MPCC is not able to find a path that goes through the gates without having to include these sharp turns and long straights. 
Even if this allows the MPCC approach to achieve higher instantaneous peak speeds (as shown in the blue color bar in Fig. \ref{fig:CPC_min_snap_gates}), the overall velocity is lower, thereby leading to suboptimal, non-competitive lap times as illustrated in Fig. \ref{fig:lap_times_sim} (labeled as \emph{Min. snap-MPCC}).}
In contrast, when the CPC path is provided as a reference, as shown in Fig. \ref{fig:lap_times_sim},  lap times (labeled as \emph{CPC-MPCC}) are very similar to the theoretical lower bound (labeled as \emph{Reference CPC}).
\change{This} shows that the MPCC is able to benefit from the long-term optimality information encoded in the CPC 3D path.
Therefore the choice of reference path is indeed highly relevant to achieve competitive lap times.
The lap times in Fig. \ref{fig:lap_times_sim} were computed \change{using the same method as} in \cite{foehn2021CPC}.
For the rest of the paper, lap times are computed in this fashion unless stated otherwise.

\rebuttal{As mentioned in section \ref{sec:pg_cpc}, and depicted in Fig. \ref{fig:solve_times_CPC_RL_PMM}, a big disadvantage of time-optimal CPC trajectories is that they are computationally demanding. This is detrimental for their use in real-time applications.
}

\subsubsection{Time-Optimal Point-Mass Model \rebuttal{(PMM)}}
\label{sec:PMM}
\begin{figure}[bt]
  \centering
  \includegraphics[width=0.95\linewidth]{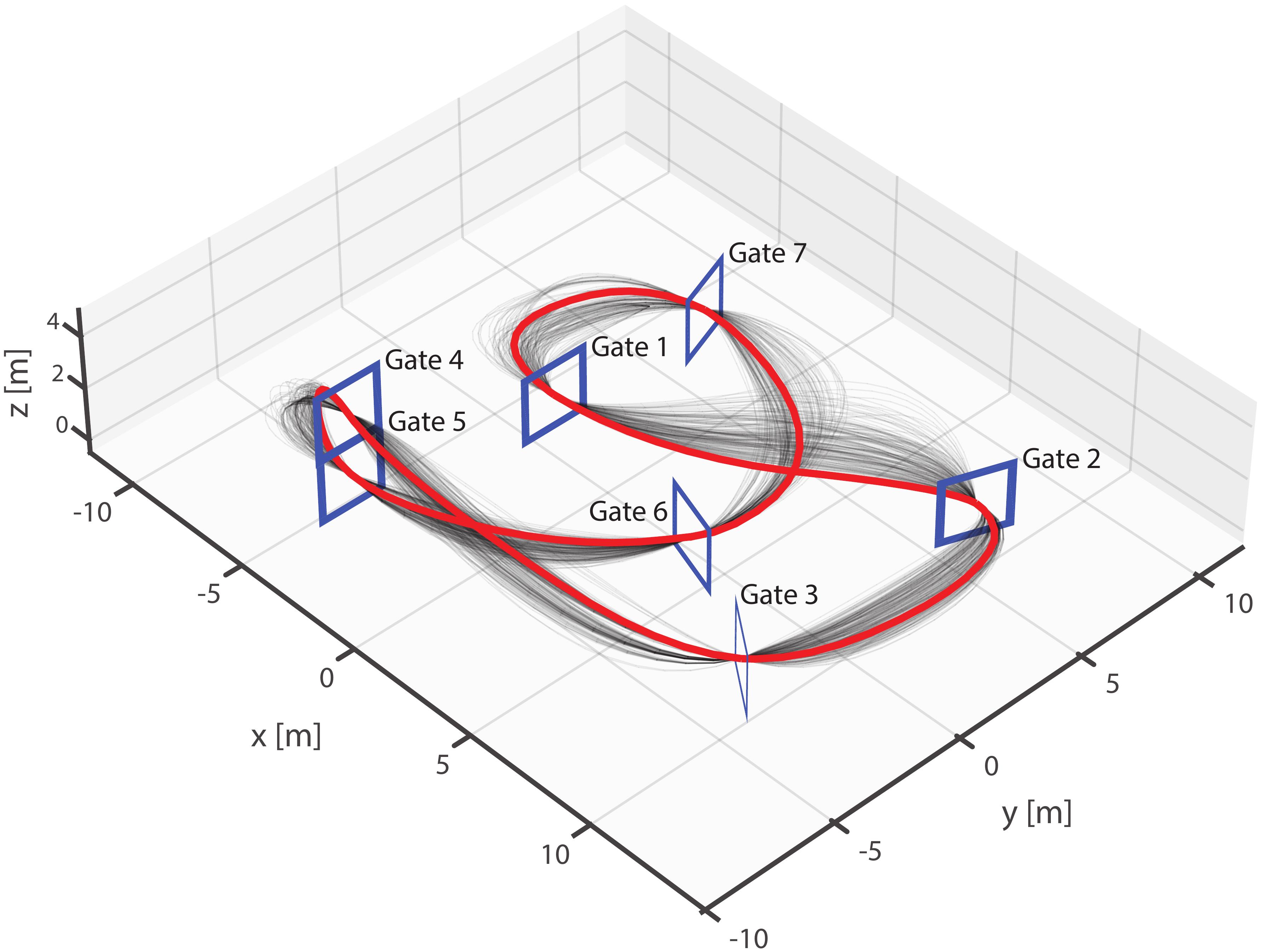}
  \caption{Time-optimal trajectory generated using a point-mass model (PMM) by sampling the velocities at the gates. The time-optimal trajectory is colored in red, while the different samples are in \change{gray}. This \change{PMM} trajectory is unfeasible for our platform, so traditional MPC methods fail to track it. However, the \change{proposed} MPCC solely needs a continuously differentiable 3D path. The computed PMM path has a shape sufficiently close to the time-optimal one that it renders \change{close to time-optimal} results when executed by our MPCC, while it can be computed in real-time.}
  \label{fig:sampling}
\end{figure}

As mentioned, one of the benefits of MPCC is that its only requirement is a continuously differentiable 3D path $\vec{p}^d(t)$.
\rebuttal{In this section we use the proposed MPCC to track time-optimal point-mass model trajectories, that do not suffer from the aforementioned computational burden.}
The resulting PMM trajectory for our race track is shown in Fig. \ref{fig:sampling}.

\begin{table}[h]
\caption{\change{Total times (averaged over 10 runs) for different gate horizons ($H_g$) for the task of completing 3 laps in the \emph{CPC-track} using the described Time-Optimal Point-Mass model (PMM) algorithm. Notice that from $H_g = 3$, the improvement in total time is negligible.}}
\label{tab:gate_ablation}
\begin{center}
    \begin{small}
    \setlength{\tabcolsep}{5pt}
        \setlength\extrarowheight{1.5pt}
        \begin{tabular}{C{1.2cm}||c|c|c}
            \toprule
            \rowcolor{gray!30!white}
            $H_g$ & Avg. Time [\si{\second}] & Min. Time [\si{\second}] & Max. Time [\si{\second}]   \\
            \hline
            1 & 16.0 & 15.75 & 16.19\\
            \hline
            2 & 15.8779 & 15.62 & 16.27 \\
            \hline
            3 &  14.8244 & 14.80 & 14.85 \\
            \hline
            4 &  14.8147 & 14.80 & 14.83 \\
            \hline
            5 &  14.818 & 14.80 & 14.83 \\
            \bottomrule
        \end{tabular}
    \end{small}
\end{center}
\end{table}

\change{In our particular application, the gate horizon \rebuttal{has been chosen to be} $H_g = 3$ gates. 
To justify this choice, we have  generated PMM trajectories for the task of completing 3 laps on the \emph{CPC-track}, for different values of $H_g$. 
We then compare the resulting total lap times (averaged over 10 runs) and show them in Table \ref{tab:gate_ablation}. 
One can notice how for $H_g$ larger than 3, the improvement in total time is negligible.
The number of samples has been chosen as in \cite{Foehn20rss}, $M = 150$ for each gate in the prediction horizon, resulting in computation times of around \SI{2}{\milli\second}.}

\begin{figure}[t]
  \centering
  \begin{subfigure}{\linewidth}
    \includegraphics[width=\linewidth]{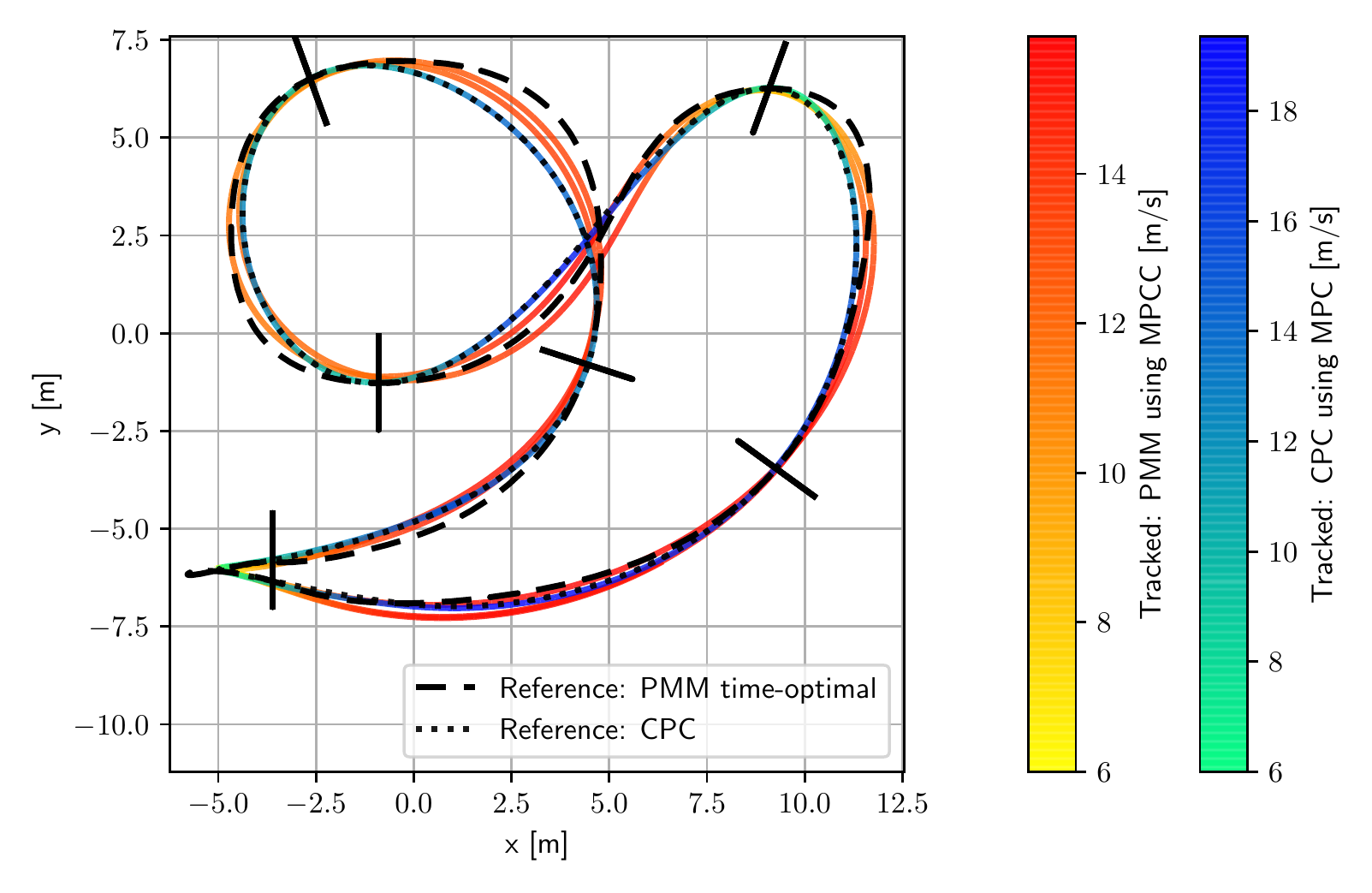}
  \end{subfigure}
  \begin{subfigure}{\linewidth}
    \includegraphics[width=0.95\linewidth]{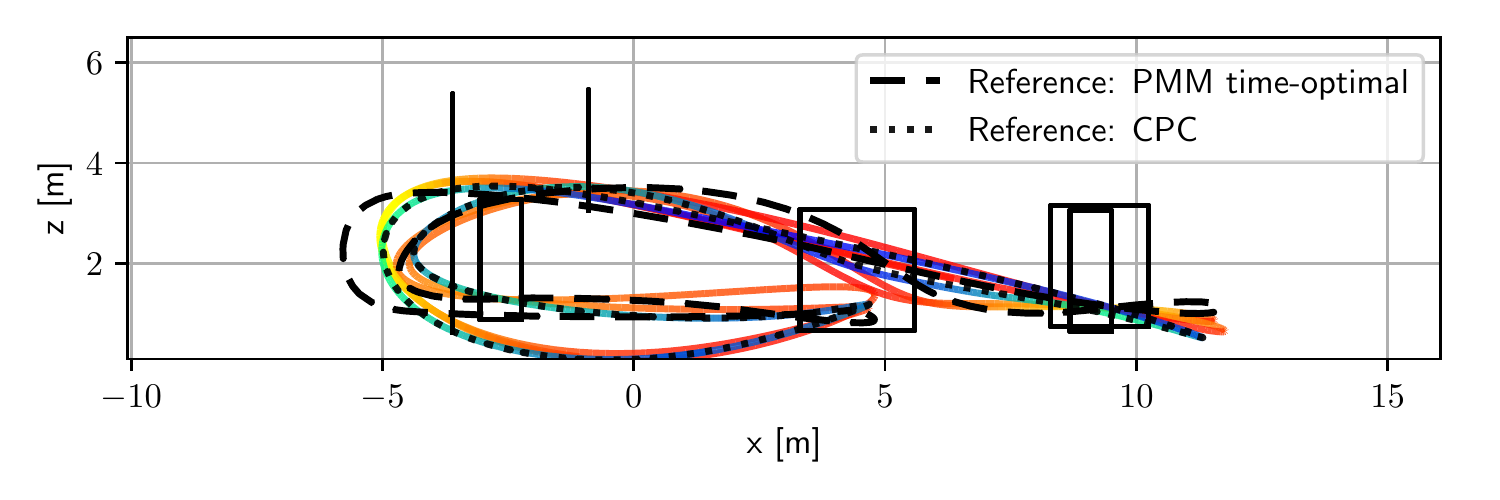}
  \end{subfigure}
  \caption{Simulated tracking performance of our MPCC approach when tracking a PMM reference (black dashed line) shown from the top view (top panel) and the side (bottom panel). The CPC time-optimal reference has been added for comparison. Notice how the position of the tracked MPCC trajectory approaches the time-optimal reference trajectory.}
  \label{fig:sim_CPC_vs_PMM}
\end{figure}

\begin{figure}[t]
  \centering
  \includegraphics[width=1.0\linewidth]{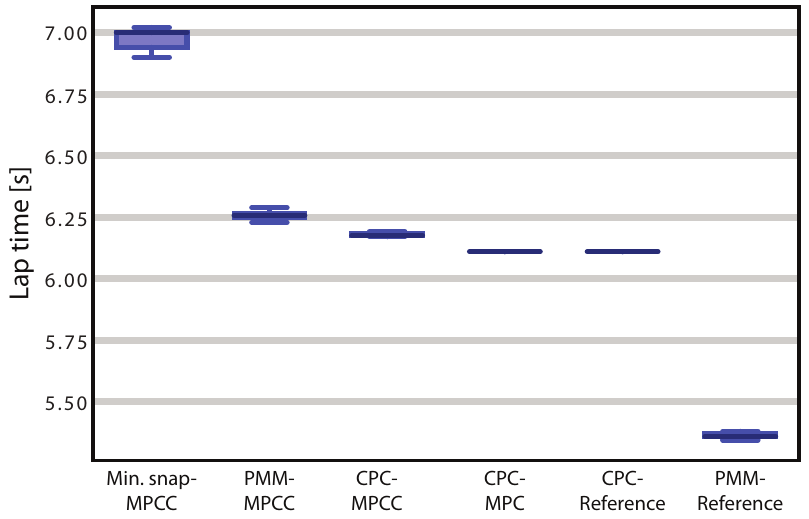}
  \caption{Sampled lap times of the different trajectories tracked using our MPCC approach in simulation and of the time-optimal reference (\emph{Reference CPC}). Notice how the point mass model tracked trajectory (\emph{PMM-MPCC}) is very close to the tracked time-optimal trajectory (\emph{CPC-MPCC}), yet does not suffer from the large computational effort required to compute a CPC trajectory. The \emph{CPC-MPC} trajectory is the CPC-Reference tracked by a standard MPC approach, for comparison. \rebuttal{Also notice how the lap times of the \emph{PMM-Reference} are lower than those of the \emph{CPC-Reference}, as expected, given that the \emph{PMM-Reference} does not satisfy the dynamics nor the constraints of a quadrotor. For completeness, the lap time of the minimum snap (reference and tracked with MPC) is 67.76 seconds, and have not been added to this figure for clarity.}}
  \label{fig:lap_times_sim}
\end{figure}

Fig. \ref{fig:sim_CPC_vs_PMM} shows the results of tracking the time-optimal point mass model trajectory with the proposed MPCC, \changeSecond{and the time-optimal CPC reference tracked with standard MPC for comparison.} 
It can be noticed that the behavior of the platform \change{approaches} the time-optimal trajectory, as also suggested by Fig. \ref{fig:hover_to_hover}, i.e., maximizing the progress along the MPCC horizon provides the short term time-optimality, whereas a \changeSecond{PMM} time-optimal reference path to track provides the controller with the long term information that \changeElia{results in} a behavior similar to a true time-optimal policy.
This can also be seen in Fig. \ref{fig:lap_times_sim}, where the lap times of the MPCC tracking a PMM reference (labeled as \emph{PMM-MPCC}) approaches the theoretical lower bound by only a small difference.

\changeSecond{In Fig. \ref{fig:lap_times_sim} one can also notice how the tracked CPC trajectory using a standard MPC approach (labeled as \emph{CPC-MPC}) has almost identical lap times as the \emph{Reference CPC}.
This is expected because standard MPC approaches use a reference that is sampled by time. 
The planned trajectory dictates at which times every reference state will be visited, meaning that the lap time is fully defined by the reference trajectory.}

%% file: sections/delay_ablation.tex
\subsection{\changeThird{Time Delay Study}}
\label{sec:delay_ablation}
\begin{figure}[t]
  \centering
  \includegraphics[width=0.95\linewidth]{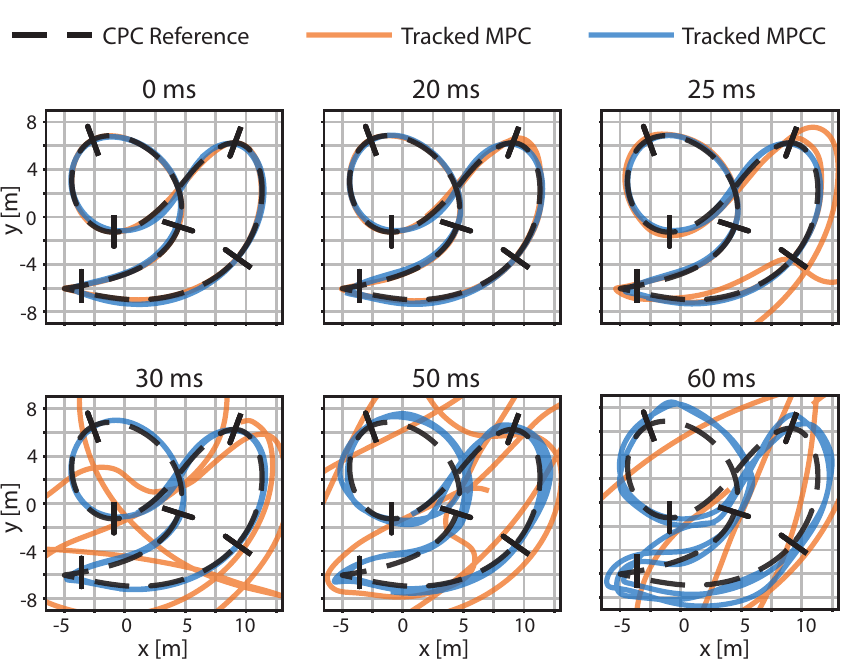}
  \caption{\changeThird{Flight performance comparison between the proposed MPCC and a standard MPC when tracking the CPC trajectory for different delays on the state estimate. The proposed MPCC can handle delays of up to \SI{50}{\milli\second} while still passing through the gates, whereas a standard MPC controller fails to track the reference at \SI{25}{\milli\second} delay.}}
  \label{fig:delay_ablation}
\end{figure}

\changeThird{Time delays are commonplace in any real system and can be originated by different sources, such as communication, sensors, or filtering algorithms, among others.
In this section, we evaluate how our controller reacts to the presence of a time delay in the control loop. 
To do so, we manually introduce a delay in the measurement of position, velocity, and attitude in our simulation.
In Fig. \ref{fig:delay_ablation} we show how both the proposed MPCC and a standard MPC react to different values of this delay, from \SI{0}{\milli\second} to \SI{60}{\milli\second}.}

\changeThird{One can notice how the proposed MPCC can successfully complete the track in the presence of up to \SI{50}{\milli\second} delay.
The explanation lies in the fact that MPCC does not use a time-sampled reference, whereas MPC does.} 
\changeThird{Standard MPC methods need a reference that consists not only of a sequence of states but also of a timestamp for each state. 
This reference is fixed a the planning stage and cannot be changed online, limiting the capabilities of the approach when a significant delay is present.
However, MPCC benefits from the liberty to select the best states at runtime and therefore can react (up to some extent) to a delay in the loop.}

%% file: sections/experiments_realworld.tex
\section{Experiment: Real-World Drone Racing}
\label{sec:exp_realworld}
\begin{figure}[t]
  \centering
  \includegraphics[width=\linewidth]{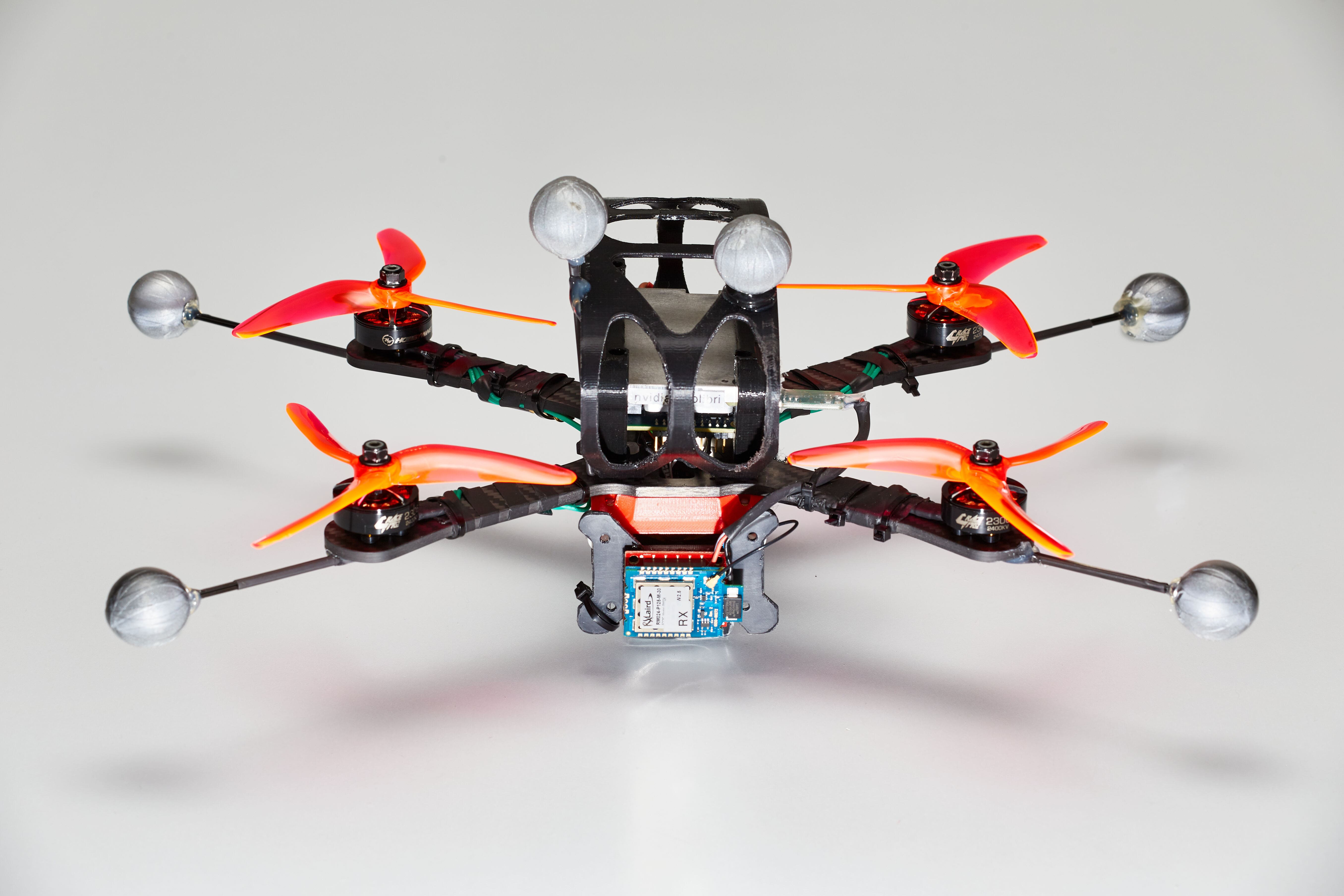}
  \caption{The quadrotor platform \rebuttal{that is} used for real-world experiments. The achievable thrust-to-weight ratio is ~4 . The platform is equipped with a Laird module for real-time wireless communication, off-the-shelf components, and reflecting markers for the VICON localization system.}
  \label{fig:drone}
\end{figure}

To further show the capabilities of the proposed MPCC for drone racing in the real world, in this section, we deploy the algorithm on the physical platform, shown in Fig. \ref{fig:drone}. 
This platform has been built in-house from off-the-shelf components and consists of an NVIDIA Jetson TX2\footnote{\url{https://developer.nvidia.com/EMBEDDED/jetson-tx2}} board that bridges the control commands via a Laird RM024\footnote{\url{https://www.lairdconnect.com/wireless-modules/ramp-ism-modules}} wireless low-latency module.  
Our method runs in an offboard desktop computer equipped with an Intel(R) Core(TM) i7-8550 CPU @ 1.80GHz.
A Radix FC board that contains the Betaflight\footnote{\url{https://betaflight.com/}} firmware is used as a low level controller. This low-level controller takes as inputs body rates and collective thrusts.
The proposed MPCC, however, uses all the dynamics of the system down to rotor thrusts. Therefore, for the real-world implementation, we use body rates and collective thrusts as inputs.
\rebuttaltwo{Due to this hardware design, the body rates are commanded directly from the MPCC controller states, and the collective thrust is computed as the sum of the single rotor thrusts.
It is important to highlight that even if in the real world experiments the inputs are the collective thrusts and body rates, these values are still generated by optimization problem \eqref{eq:full_ocp}, and therefore they are dynamically feasible and they respect the single rotor thrust constraints.
This way, the body rates and collective thrust are generated taking into account possible current and predicted saturations at the single rotor thrust level. For this reason, it is still advantageous that the MPCC considers the full state dynamics. 
}

For state estimation, we use a VICON\footnote{\url{https://www.vicon.com/}} system with 36 cameras \change{in one of the world's largest drone flying arenas ($30\times 30\times 8$\,m)} that provide the platform with down to millimeter accuracy measurements of position and orientation. 

\subsection{Implementation Details}
In order to \change{deploy} our MPCC controller, problem \eqref{eq:full_ocp} needs to be solved in real-time. To this end, we use the ACADO\footnote{\url{https://acado.github.io/}} toolkit as a code generation tool, together with QPOASES\footnote{\url{https://projects.coin-or.org/qpOASES}} as a QP solver. The optimization problem is solved following a real-time iteration scheme \cite{Diehl2006springer} with a feedback rate of $100$ Hz and a prediction rate of $16.6$ Hz.

\begin{figure}[t]
  \centering
  \includegraphics[width=0.95\linewidth]{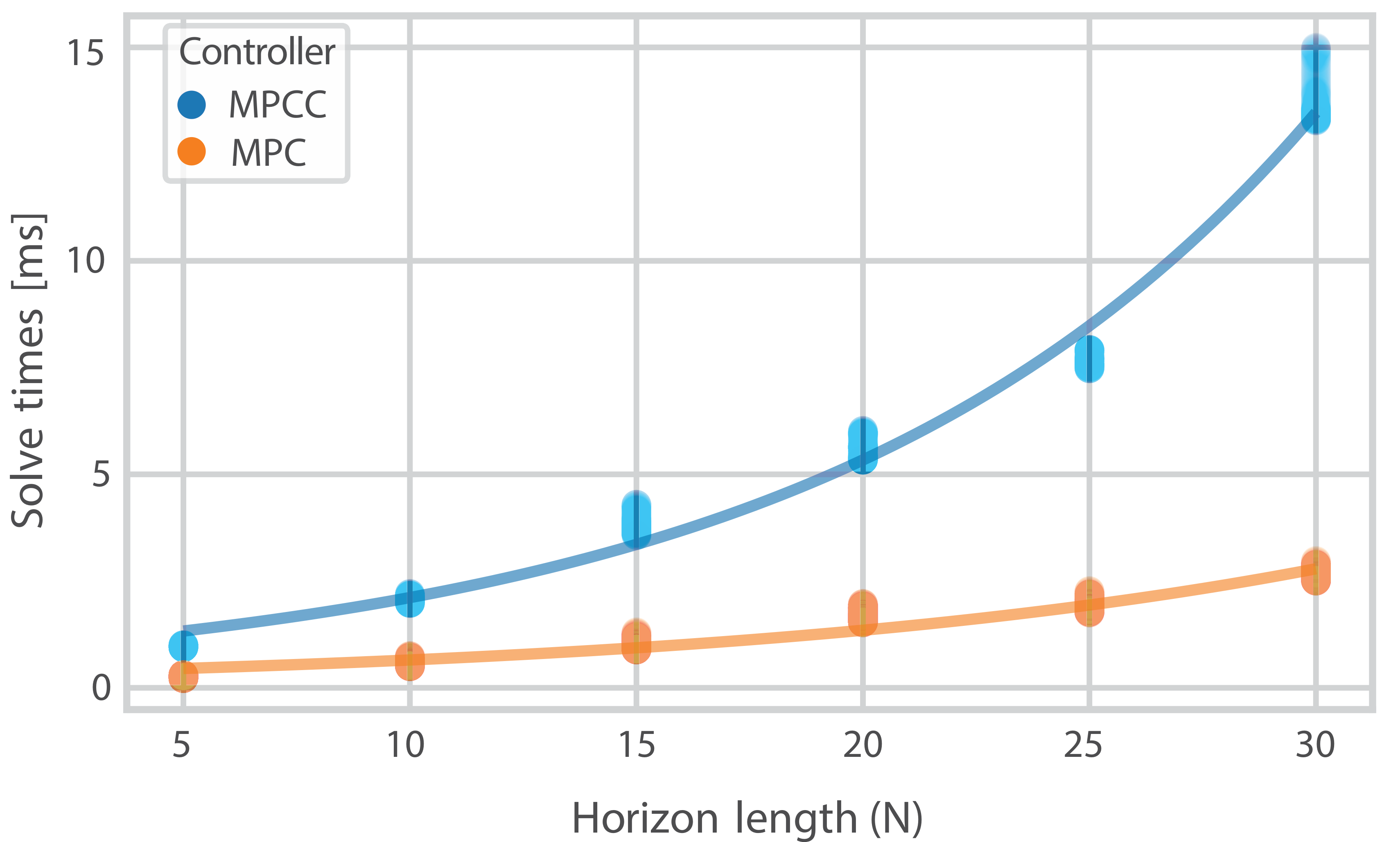}
  \caption{\changeFourth{Evolution of solver timings of MPCC (optimization problem \eqref{eq:full_ocp}) and MPC (as in \cite{foehn2021CPC}), with respect to horizon length (N). The optimization problems are solved on an desktop computer equipped with Intel(R) Core(TM) i7-8550 CPU @ 1.80GHz.}}
  \label{fig:boxplot_solvetimes}
\end{figure}

\subsection{\change{Solver Timings}}

Fig. \ref{fig:boxplot_solvetimes} shows the evolution of the \change{solver timings} for different horizon lengths, \changeFourth{for both MPCC (optimization problem \eqref{eq:full_ocp}) and a standard MPC (as in \cite{foehn2021CPC})}. 
For MPCC to be able to call the solver every $10 \; ms$, the maximum horizon length would be $N = 25$. However, that would leave the system with limited computation time for other modules that also need to run during the remaining time, such as the spline computation \changeElia{(explained in Section \ref{sec:arc_length_splines})}.
In order to ensure that the entire control pipeline runs smoothly at all times, we have selected a horizon length of $N = 20$ for all our MPCC experiments in both simulation and the real world. This gives a solver time of about $5 \; ms$.

\subsection{Baselines}
\label{sec:baselines}

To benchmark \change{the performance of our approach} in real-world flight, we compare it to (i) the fastest executed CPC trajectory in \cite{foehn2021CPC} and (ii) a world-class professional drone racing pilot (\changeFourth{Timothy Trowbridge, whose list of achievements was previously reported in \cite{foehn2021CPC}}).

\changeThird{It is important to remark that by definition, CPC trajectories take full advantage of the available actuator power at all times. 
Therefore they leave no actuation power for control authority under model mismatches and disturbances, and even the smallest deviation could potentially lead to a crash.
While this might not be an issue in simulated environments, it definitely is in real-world flights.
Because of this, in order to provide robustness and add a control authority safety margin, for the real flight experiments in \cite{foehn2021CPC} the authors planned the trajectory using 3.3 as thrust-to-weight ratio (TWR) but  executed it with slightly more thrust, of around 4 TWR. 
To have a faithful comparison, we use exactly the same platform with the same limits in our experiments.}

\change{The platform used by the professional drone pilot is very similar to the one used to deploy our algorithm (shown in Fig. \ref{fig:drone})}, where the Jetson TX2 and the Laird modules were substituted by an FPV camera system and an RC receiver. Reflective markers were also attached to the human platform to allow precise data collection. The TWR of this platform was matched to be the same as the one used for our real-world experiments.

\subsection{Results}
A video of the experimental results of this work can be found in \changeFifth{\url{https://youtu.be/mHDQcckqdg4}}. These experiments serve as proof of concept of the real-world applicability of the proposed controller. \\

\begin{figure*}[tp]
  \centering
  \includegraphics[width=0.95\textwidth]{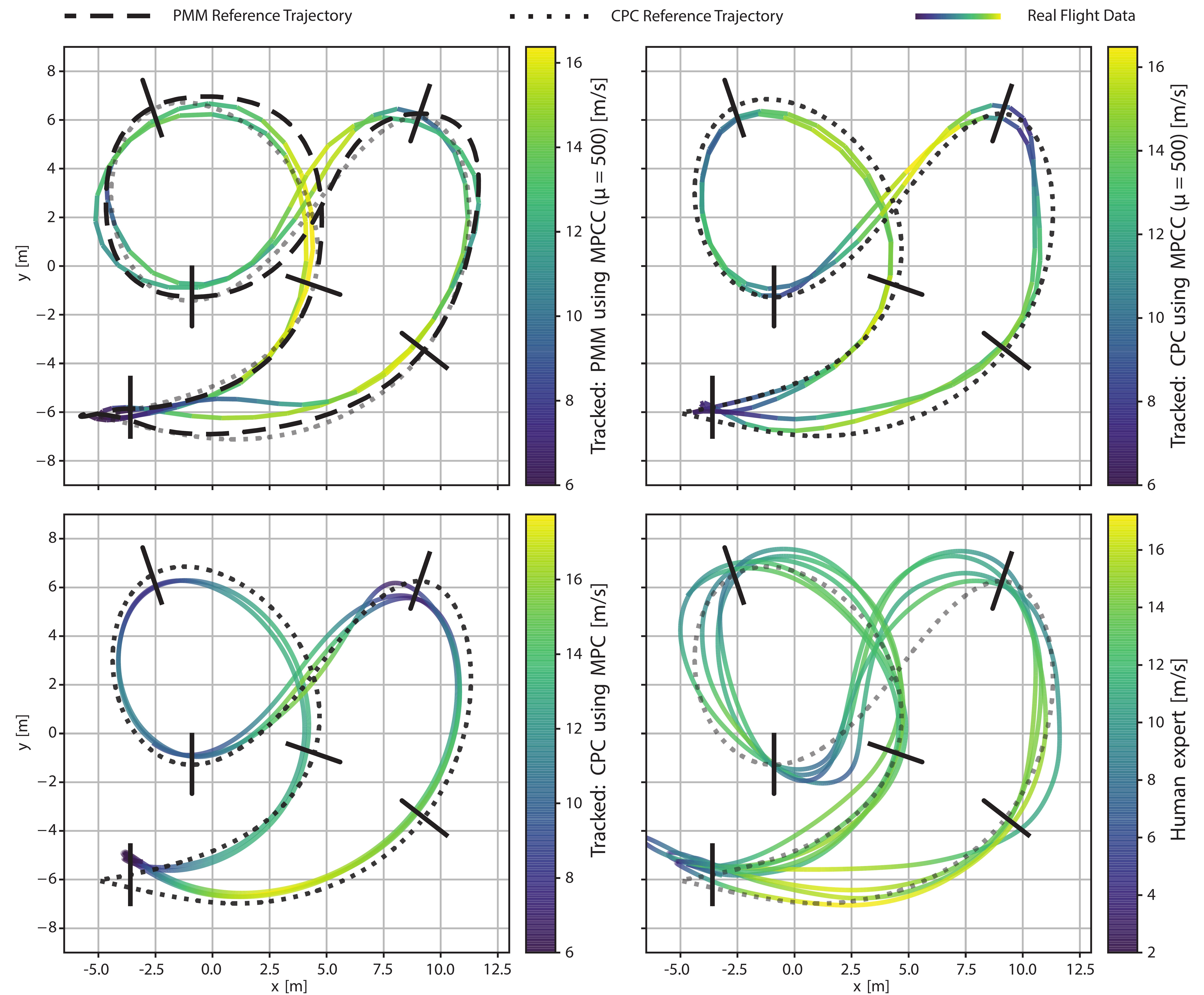}
  \caption{\changeFourth{Real world experiment of the MPCC tracking a time-optimal trajectory generated using a point mass model (PMM Reference, top left), and of the MPCC tracking a true time-optimal trajectory (CPC Reference, top right), both using $\mu = 500$. We compare our results with the executed time-optimal trajectory (CPC tracked by MPC, bottom left), and with the best human expert from \cite{foehn2021CPC} (bottom right). Note how when using MPCC as a controller, the platform stays for longer times at high speeds regime.}}
  \label{fig:exp_CPC_vs_PMM}
\end{figure*}

\changeFourth{Fig. \ref{fig:exp_CPC_vs_PMM} shows our results in the top row and the baselines in the bottom row. 
From left to right and top to bottom, we show the real flight data from our MPCC tracking the PMM reference, our MPCC tracking the CPC reference, a standard MPC tracking the CPC reference (as in \cite{foehn2021CPC}), and a world-class professional human expert.
The CPC time-optimal reference has been added with transparency to the PMM-MPCC plot and to the human expert plot for comparison purposes only.
It is interesting to notice how the PMM-MPCC executed trajectory approaches the theoretical full model time-optimal (labeled as \emph{CPC reference}), as already suggested in the simulation experiments (Fig. \ref{fig:sim_CPC_vs_PMM}).}

\begin{figure}[t]
  \centering
  \includegraphics[width=\linewidth]{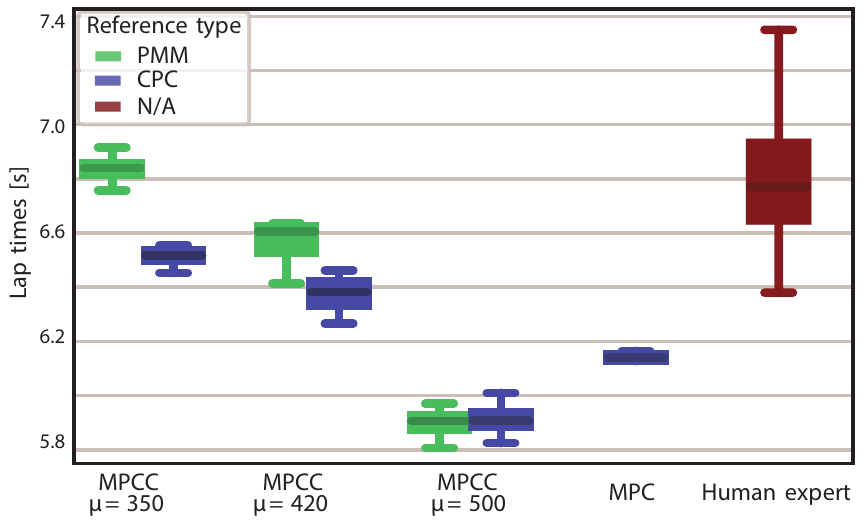}
  \caption{\changeThird{Lap times of our MPCC approach tracking both the PMM and the CPC references, for different values of progress cost $\mu$. We compare it to the execution of the CPC trajectory by a standard MPC controller and to a human expert flying the same track.}}
  \label{fig:boxplot_laptimes}
\end{figure}

\changeFourth{Fig. \ref{fig:boxplot_laptimes} shows the lap time distribution of our MPCC approach (tracking both CPC and PMM references) compared to those of the standard MPC tracking a CPC reference and the human expert. 
As an ablation study, we show how our approach performs for different values of $\mu$. 
As shown by the original MPCC formulation in \eqref{eq:ocp_generic}, the progress weight contributes mainly to making the trajectory go faster by trading it off for contour error.
In this same figure, we notice that when using the CPC path as reference to the MPCC controller, the lap times are faster than when using the PMM as reference, in concordance with the results and conclusions from the ablation study performed in Section \ref{sec:path_generation_drone}.
In our experiments for larger progress costs than $\mu = 500$, the error around the gates starts being significantly larger, resulting in the platform passing too far from the gates and leading to crashes against the gate frames.}
 
\begin{figure}[t]
  \centering
  \includegraphics[width=\linewidth]{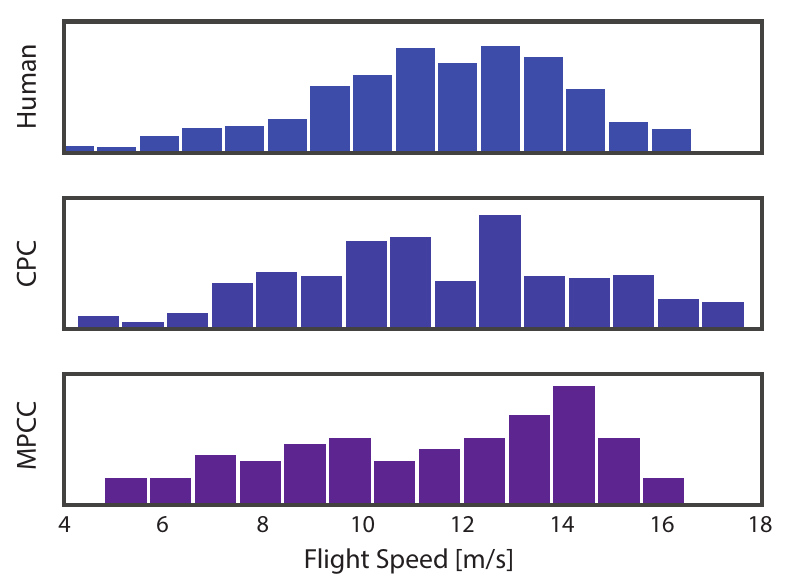}
  \caption{\changeFourth{Histogram of flight speed of different control approaches. From top to bottom: human expert trajectory, CPC trajectory tracked with standard MPC, and CPC trajectory tracked with the proposed MPCC. Note how the MPCC controller spends more time at higher speeds.}}
  \label{fig:hist_flight_speed}
\end{figure}

\changeThird{When comparing our MPCC method to the CPC tracked by MPC, we observe that even though the tracked MPCC trajectory has a slightly lower maximum velocity (color bars in Fig. \ref{fig:exp_CPC_vs_PMM}), the lap times are better for our MPCC approach (see Fig. \ref{fig:boxplot_laptimes}, \emph{MPCC $\mu = 500$} vs. \emph{MPC}). 
This can be explained by the fact that our MPCC approach can effectively take better advantage of the control authority safety margin (already introduced in Section \ref{sec:baselines}). 
The proposed MPCC controller benefits from the freedom to decide the future states of the platform online, meaning that it is able to adapt and command full thrust when possible or re-gain control authority when necessary.
\rebuttal{This means that in the presence of disturbances or deviations with respect to the nominal path, the MPCC controller generates at every iteration a new time allocation and a new sequence of states and inputs that takes into account its current state and the actuation limits, while at the same time maximizing the progress.}
Consequently, the proposed MPCC approach results in higher speeds for longer periods of time, as can be noted in the colored profile in Fig. \ref{fig:exp_CPC_vs_PMM} and in the flight speed histogram in Fig. \ref{fig:hist_flight_speed}.
In contrast, the MPC approach is based on minimizing the error with respect to the pre-computed reference,
\rebuttal{
which is generated offline, and only once.
In the presence of disturbances from the reference trajectory, while actuators are close to saturation, there is no control authority left to correct these deviations. Because \rebuttal{the CPC trajectory} generation is very computationally expensive, it is not possible to promptly generate a new sequence of times, states, and inputs that consider the current state of the platform.
As previously mentioned in Section \ref{sec:baselines}, this is the reason why we need to plan the CPC trajectories considering a limit that is lower than the actual platform limit, such that in the event of a deviation from the reference, there is enough control authority left to correct for it.
}
As a side note, these effects are not noticeable in our simulated environments (Section \ref{sec:path_generation_drone}) because there are no major disturbances in our simulation.
Therefore, there is no need to add a control authority safety margin, meaning that the \emph{MPC} approach is able to track the trajectory with the same \rebuttal{actuation limits} with which it was planned.}

\changeFourth{Additionally, as shown in Section \ref{sec:delay_ablation}, the proposed MPCC approach is more robust at handling delays than standard MPC.
Since all real systems are prone to have delays (in our experiments, mainly coming from the communication involving the motion capture system and the LAIRD module), this also potentially contributes to the advantage of our MPCC approach.
}

When comparing the autonomously flown trajectories to the human expert, one can notice how the human trajectories are opening more and overshooting in most turns. This is related to the vision-based first-person view flying behavior since human pilots  need to orient the camera toward the upcoming waypoints by applying coordinated yaw, pitch, and roll commands. Moreover, our recent work indicates that human pilots use long prediction horizons related to visual-motor response latencies in the order of 220 ms \cite{Pfeiffer_Scaramuzza_2021}.

%% file: sections/discussion.tex
\section{Discussion}
\label{sec:discussion}
\subsection{Convexity}
The convexity of the cost function \eqref{eq:J_mpcc} depends solely on the convexity of $\Vert e_{c,k} \Vert_{q_c}^2$, since $\rho \theta_N$ is an affine term of an optimization variable and therefore does not affect convexity.
It follows that optimization problem \eqref{eq:full_ocp} is non-convex due to (i) as $p^d(\theta_k)$ is formulated as a set of third-order splines, $\vec{e}(\theta_k)$ is also a third-order polynomial, and $\vec{t}(\theta_k)$ is of second order.
Hence, the norm of the contour \eqref{eq:qc_norm} and lag \eqref{eq:ql_norm} errors result in polynomials of degrees 24 and 12, respectively. And (ii) the rotational dynamics of the system \eqref{eq:quad_dynamics} are non-linear.
Consequently, problem \eqref{eq:full_ocp} is a Non-Linear Program (NLP), and different strategies can be taken to solve it in real-time. 
In the literature, this is commonly approached by manually carrying out the linearization of the errors and the dynamics around the previous prediction, effectively converting the NLP into a Quadratic Program (QP) \cite{Liniger_Domahidi_Morari_2015}, \cite{Lam_Manzie_Good_2010}. 
In other approaches, an interior point method solver is chosen to solve the NLP directly \cite{Schwarting_Alonso-Mora_Paull_Karaman_Rus_2018}. 
In our work, the optimization problem follows an RTI scheme \cite{Diehl2006springer}. The non-linear problem is directly fed to ACADO, which carries out a linearization at every time step to convert the NLP into a QP (preparation phase) that is then solved by QPOASES (feedback phase).
\subsection{Tuning}
Controller tuning is one of the main challenges among the control community. The sensitivity to control tuning is particularly pronounced when talking about aggressive, high-speed quadrotor flight \cite{autotune_Loquercio_Saviolo_Scaramuzza_2022}. 
Although in the cost function from problem \eqref{eq:full_ocp} there are many different weights, most of these parameters act as regularization terms ($\bm{Q_w}, r_{\Delta v}, \bm{R_{\Delta f}}$) not to allow the controller to choose arbitrarily big control inputs; and $q_l$, which is chosen big enough such that the error approximations are sufficiently small. 
Once these terms are tuned, they are kept constant for all different conditions in both simulation and real-world experiments. 

That leaves only two different scalar values to tune, $q_c$ and $\mu$. 
Therefore, the benefits of using MPCC as control strategy in comparison to standard methods are not only related to the liberty the controller offers by choosing the best sampling times and states but also that it significantly reduces the dimensionality of the hyperparameter space.
However, it is important to note that, because of the dynamic allocation of the weights performed when using our method (described in section \ref{sec:dyn_weight}), the number of different Gaussians associated to $q_c$ depends on the number of waypoints. This can potentially be alleviated by using learning approaches to find the optimal tuning parameters given a higher level objective. For instance, reinforcement-learning--based approaches have shown to be promising in previous work \cite{Song_Scaramuzza}.

\changeFourth{Another advantage of MPCC is that the speed at which the platform tracks the path can be selected by changing the progress weight $\mu$ accordingly. In contrast, for a standard reference tracking control method (such as MPC), if the desired speed of the platform needs to be changed, the entire reference needs to be computed again.}


\subsection{Limitations}
Albeit the proposed controller has shown to achieve outstanding performance in highly agile quadrotor flight, it currently runs on an external desktop PC (Intel(R) Core(TM) i7-8550 CPU @ 1.80GHz) at $5 \; ms$.
Since the optimization problem \eqref{eq:full_ocp} to be solved in each iteration is quite complex, it is currently difficult to run it onboard embedded computers like the NVIDIA Jetson TX2.
This could be tackled by using new state-of-the-art solvers specifically designed for Model Predictive Control formulations that have shown to outperform QPOASES \cite{HPIPM}.

\section{Conclusion}
This paper proposes a Model Predictive Contouring Control formulation in the framework of full quadcopter dynamics. 
We highlight and exploit the main advantages of this formulation and show that it is very well suited for the problem of drone racing. 
In particular, we use a sampling-based method to generate non-feasible, close-to-time-optimal trajectories that can then be tracked with our controller.
Additionally, we encode the gates in the optimization problem by adapting the contouring weights in the cost function depending on the gates' location. 
We compare our controller to a standard MPC tracking full state time-optimal trajectories and show that they render superhuman performance in both simulation and real-world experiments, with the advantage that our approach could be re-planned in real-time.
\changeFourth{Additionally, we show that, when deployed in the real platform, our MPCC approach is able to take better advantage of the full actuator potential than a standard MPC approach tracking a time-optimal trajectory, resulting in better lap times.}

The results from this work bring the field of autonomous drone racing closer to beating the best humans in a fair drone race.
Although there are still challenges ahead mainly related to onboard state estimation and perception at high speeds, the proposed controller has proven to be a stable approach to fly very aggressive maneuvers at very high speeds.

Finally, the control method presented in this work can find utility in other applications.
Specifically, the fact that our method can track non-feasible trajectories and that it allows the selection of desired speeds \changeElia{reduces} the required computation effort needed for planning. 
This opens the door to, for example, several sampling-based approaches where the dynamic feasibility constraint could be dropped.

\rebuttalthree{
\section{Acknowledgments}
The authors want to thank Christian Pfeiffer for his help with gathering the professional human pilots' data and Thomas Laengle for his support with the experimental setup.
}

%% file: main.bbl
\begin{thebibliography}{10}
\providecommand{\url}[1]{#1}
\csname url@samestyle\endcsname
\providecommand{\newblock}{\relax}
\providecommand{\bibinfo}[2]{#2}
\providecommand{\BIBentrySTDinterwordspacing}{\spaceskip=0pt\relax}
\providecommand{\BIBentryALTinterwordstretchfactor}{4}
\providecommand{\BIBentryALTinterwordspacing}{\spaceskip=\fontdimen2\font plus
\BIBentryALTinterwordstretchfactor\fontdimen3\font minus
  \fontdimen4\font\relax}
\providecommand{\BIBforeignlanguage}[2]{{%
\expandafter\ifx\csname l@#1\endcsname\relax
\typeout{** WARNING: IEEEtran.bst: No hyphenation pattern has been}%
\typeout{** loaded for the language `#1'. Using the pattern for}%
\typeout{** the default language instead.}%
\else
\language=\csname l@#1\endcsname
\fi
#2}}
\providecommand{\BIBdecl}{\relax}
\BIBdecl

\bibitem{Loianno2020jfr}
G.~Loianno and D.~Scaramuzza, ``Special issue on future challenges and
  opportunities in vision-based drone navigation,'' \emph{J. Field Robot.},
  2020.

\bibitem{air_taxi_2020}
\BIBentryALTinterwordspacing
S.~Rajendran and S.~Srinivas, ``Air taxi service for urban mobility: a critical
  review of recent developments, future challenges, and opportunities,''
  \emph{Transportation Research Part E-logistics and Transportation Review},
  Nov 2020. [Online]. Available: \url{https://www.scinapse.io}
\BIBentrySTDinterwordspacing

\bibitem{Mellinger12ijrr}
D.~Mellinger, N.~Michael, and V.~Kumar, ``Trajectory generation and control for
  precise aggressive maneuvers with quadrotors,'' \emph{Int. J. Robot.
  Research}, 2012.

\bibitem{loianno2016estimation}
G.~Loianno, C.~Brunner, G.~McGrath, and V.~Kumar, ``Estimation, control, and
  planning for aggressive flight with a small quadrotor with a single camera
  and imu,'' \emph{IEEE Robotics and Automation Letters}, vol.~2, no.~2, pp.
  404--411, 2017.

\bibitem{kaufmann18icra}
E.~Kaufmann, M.~Gehrig, P.~Foehn, R.~Ranftl, A.~Dosovitskiy, V.~Koltun, and
  D.~Scaramuzza, ``Beauty and the beast: Optimal methods meet learning for
  drone racing,'' \emph{{IEEE} Int. Conf. Robot. Autom. (ICRA)}, pp. 690--696,
  2018.

\bibitem{Mohta18jfr}
K.~Mohta, M.~Watterson, Y.~Mulgaonkar, S.~Liu, C.~Qu, A.~Makineni, K.~Saulnier,
  K.~Sun, A.~Zhu, J.~Delmerico, K.~Karydis, N.~Atanasov, G.~Loianno,
  D.~Scaramuzza, K.~Daniilidis, C.~J. Taylor, and V.~Kumar, ``Fast, autonomous
  flight in gps-denied and cluttered environments,'' \emph{J. Field Robot.},
  vol.~35, no.~1, pp. 101--120, 2018.

\bibitem{Zhou19TRO}
B.~Zhou, J.~Pan, F.~Gao, and S.~Shen, ``{RAPTOR:} robust and perception-aware
  trajectory replanning for quadrotor fast flight,'' \emph{IEEE Transactions on
  Robotics}, 2021.

\bibitem{Foehn20rss}
\BIBentryALTinterwordspacing
P.~Foehn, D.~Brescianini, E.~Kaufmann, T.~Cieslewski, M.~Gehrig, M.~Muglikar,
  and D.~Scaramuzza, ``Alphapilot: Autonomous drone racing,'' \emph{Robotics:
  Science and Systems (RSS)}, 2020. [Online]. Available:
  \url{https://link.springer.com/article/10.1007/s11370-018-00271-6}
\BIBentrySTDinterwordspacing

\bibitem{croonRAS20}
S.~Li, M.~M. Ozo, C.~D. Wagter, and G.~C. de~Croon, ``Autonomous drone race: A
  computationally efficient vision-based navigation and control strategy,''
  \emph{Robotics and Autonomous Systems}, vol. 133, 2020.

\bibitem{MPCSurveyASL_2020}
H.~Nguyen, M.~Kamel, K.~Alexis, and R.~Siegwart, ``Model predictive control for
  micro aerial vehicles: A survey,'' \emph{2021 European Control Conference
  (ECC)}, 2021.

\bibitem{DDAKaufmann_Loquercio_Scaramuzza_2020}
\BIBentryALTinterwordspacing
E.~Kaufmann, A.~Loquercio, R.~Ranftl, M.~Müller, V.~Koltun, and D.~Scaramuzza,
  ``Deep drone acrobatics,'' in \emph{Robotics: Science and Systems XVI}.\hskip
  1em plus 0.5em minus 0.4em\relax Robotics: Science and Systems Foundation,
  Jul 2020. [Online]. Available:
  \url{http://www.roboticsproceedings.org/rss16/p040.pdf}
\BIBentrySTDinterwordspacing

\bibitem{foehn2021CPC}
\BIBentryALTinterwordspacing
P.~Foehn, A.~Romero, and D.~Scaramuzza, ``Time-optimal planning for quadrotor
  waypoint flight,'' \emph{Science Robotics}, vol.~6, no.~56, 2021. [Online].
  Available: \url{https://robotics.sciencemag.org/content/6/56/eabh1221}
\BIBentrySTDinterwordspacing

\bibitem{Moon19jirc}
H.~Moon, J.~{Martinez-Carranza}, T.~Cieslewski, M.~Faessler, D.~Falanga,
  A.~Simovic, D.~Scaramuzza, S.~Li, M.~Ozo, C.~{De Wagter}, G.~{de Croon},
  S.~Hwang, S.~Jung, H.~Shim, H.~Kim, M.~Park, T.~Au, and S.~J. Kim,
  ``Challenges and implemented technologies used in autonomous drone racing,''
  \emph{J. Intell. Service Robot.}, 2019.

\bibitem{cocoma2019towards}
J.~A. Cocoma-Ortega and J.~Mart{\'\i}nez-Carranza, ``Towards high-speed
  localisation for autonomous drone racing,'' in \emph{Mexican International
  Conference on Artificial Intelligence}.\hskip 1em plus 0.5em minus
  0.4em\relax Springer, 2019.

\bibitem{Madaan20arxiv}
R.~Madaan, N.~Gyde, S.~Vemprala, M.~Brown, K.~Nagami, T.~Taubner,
  E.~Cristofalo, D.~Scaramuzza, M.~Schwager, and A.~Kapoor, ``Airsim drone
  racing lab,'' \emph{PMLR post-proceedings of the NeurIPS 2019's Competition
  Track}, 2020.

\bibitem{guerra2019flightgoggles}
W.~Guerra, E.~Tal, V.~Murali, G.~Ryou, and S.~Karaman, ``{FlightGoggles}:
  Photorealistic sensor simulation for perception-driven robotics using
  photogrammetry and virtual reality,'' in \emph{{IEEE}/{RSJ} International
  Conference on Intelligent Robots and Systems ({IROS})}.\hskip 1em plus 0.5em
  minus 0.4em\relax {IEEE}, 2019.

\bibitem{AgileFlight}
{CORDIS - European Comission}, ``{AgileFlight},''
  \url{https://cordis.europa.eu/project/id/864042}, accessed: 2021-7-30.

\bibitem{blakbox_karaman_2020}
\BIBentryALTinterwordspacing
G.~Ryou, E.~Tal, and S.~Karaman, ``Multi-fidelity black-box optimization for
  time-optimal quadrotor maneuvers,'' in \emph{Robotics: Science and Systems
  XVI}.\hskip 1em plus 0.5em minus 0.4em\relax Robotics: Science and Systems
  Foundation, Jul 2020. [Online]. Available:
  \url{http://www.roboticsproceedings.org/rss16/p032.pdf}
\BIBentrySTDinterwordspacing

\bibitem{Maurer}
H.~Maurer, ``On optimal control problems with bounded state variables and
  control appearing linearly,'' \emph{SIAM Journal on Control and
  Optimization}, vol.~15, no.~3, pp. 345--362, 1977.

\bibitem{Lam_Manzie_Good_2010}
D.~Lam, C.~Manzie, and M.~Good, ``Model predictive contouring control,'' in
  \emph{49th IEEE Conference on Decision and Control (CDC)}, Dec 2010, p.
  6137–6142.

\bibitem{Liniger_Domahidi_Morari_2015}
A.~Liniger, A.~Domahidi, and M.~Morari,
  ``\BIBforeignlanguage{en}{Optimization-based autonomous racing of 1:43 scale
  rc cars},'' \emph{\BIBforeignlanguage{en}{Optimal Control Applications and
  Methods}}, vol.~36, no.~5, p. 628–647, 2015.

\bibitem{lee2010geometric}
T.~Lee, M.~Leok, and N.~H. McClamroch, ``Geometric tracking control of a
  quadrotor uav on se (3),'' in \emph{49th IEEE conference on decision and
  control (CDC)}.\hskip 1em plus 0.5em minus 0.4em\relax IEEE, 2010, pp.
  5420--5425.

\bibitem{fresk2013full}
E.~Fresk and G.~Nikolakopoulos, ``Full quaternion based attitude control for a
  quadrotor,'' in \emph{2013 European control conference (ECC)}.\hskip 1em plus
  0.5em minus 0.4em\relax IEEE, 2013, pp. 3864--3869.

\bibitem{brescianini2018tilt}
D.~Brescianini and R.~D’Andrea, ``Tilt-prioritized quadrocopter attitude
  control,'' \emph{IEEE Transactions on Control Systems Technology}, vol.~28,
  no.~2, pp. 376--387, 2018.

\bibitem{Bangura14ifac}
M.~Bangura and R.~Mahony, ``Real-time model predictive control for
  quadrotors,'' \emph{{IFAC} World Congress}, 2014.

\bibitem{Diehl2006springer}
M.~Diehl, H.~G. Bock, H.~Diedam, and P.~B. Wieber, ``Fast direct multiple
  shooting algorithms for optimal robot control,'' in \emph{Fast motions in
  biomechanics and robotics}.\hskip 1em plus 0.5em minus 0.4em\relax Springer,
  2006.

\bibitem{svacha2017improving}
J.~Svacha, K.~Mohta, and V.~Kumar, ``Improving quadrotor trajectory tracking by
  compensating for aerodynamic effects,'' in \emph{2017 international
  conference on unmanned aircraft systems (ICUAS)}.\hskip 1em plus 0.5em minus
  0.4em\relax IEEE, 2017, pp. 860--866.

\bibitem{torrente2021data}
G.~Torrente, E.~Kaufmann, P.~F{\"o}hn, and D.~Scaramuzza, ``Data-driven mpc for
  quadrotors,'' \emph{IEEE Robotics and Automation Letters}, vol.~6, no.~2, pp.
  3769--3776, 2021.

\bibitem{tal2020accurate}
E.~Tal and S.~Karaman, ``Accurate tracking of aggressive quadrotor trajectories
  using incremental nonlinear dynamic inversion and differential flatness,''
  \emph{IEEE Transactions on Control Systems Technology}, 2020.

\bibitem{Mellinger11icra}
D.~Mellinger and V.~Kumar, ``Minimum snap trajectory generation and control for
  quadrotors,'' in \emph{{IEEE} Int. Conf. Robot. Autom. (ICRA)}, 2011.

\bibitem{Foehn17rss}
P.~Foehn, D.~Falanga, N.~Kuppuswamy, R.~Tedrake, and D.~Scaramuzza, ``Fast
  trajectory optimization for agile quadrotor maneuvers with a cable-suspended
  payload,'' in \emph{Robotics: Science and Systems (RSS)}, 2017.

\bibitem{Hehn12ar}
M.~Hehn, R.~Ritz, and R.~D'Andrea, ``Performance benchmarking of quadrotor
  systems using time-optimal control,'' \emph{Auton. Robots}, Mar. 2012.

\bibitem{Loock13ecc}
W.~V. Loock, G.~Pipeleers, and J.~Swevers, ``Time-optimal quadrotor flight,''
  in \emph{{IEEE} Eur. Control Conf. (ECC)}, 2013.

\bibitem{spedicato2017minimum}
S.~Spedicato and G.~Notarstefano, ``Minimum-time trajectory generation for
  quadrotors in constrained environments,'' \emph{IEEE Transactions on Control
  Systems Technology}, vol.~26, no.~4, pp. 1335--1344, 2017.

\bibitem{Allen16gnc}
R.~Allen and M.~Pavone, \emph{A Real-Time Framework for Kinodynamic Planning
  with Application to Quadrotor Obstacle Avoidance}, 2016.

\bibitem{sampling_lq_mt_Kumar_2017}
S.~Liu, N.~Atanasov, K.~Mohta, and V.~Kumar, ``Search-based motion planning for
  quadrotors using linear quadratic minimum time control,'' in \emph{2017
  IEEE/RSJ International Conference on Intelligent Robots and Systems (IROS)},
  Sep 2017, p. 2872–2879.

\bibitem{sampling_SE3_Kumar_2018}
S.~Liu, K.~Mohta, N.~Atanasov, and V.~Kumar, ``Search-based motion planning for
  aggressive flight in se(3),'' \emph{IEEE Robotics and Automation Letters},
  vol.~3, no.~3, p. 2439–2446, Jul 2018.

\bibitem{jorris2009three}
T.~R. Jorris and R.~G. Cobb, ``Three-dimensional trajectory optimization
  satisfying waypoint and no-fly zone constraints,'' \emph{Journal of Guidance,
  Control, and Dynamics}, vol.~32, no.~2, pp. 551--572, 2009.

\bibitem{bousson20134d}
K.~Bousson and P.~F. Machado, ``4d trajectory generation and tracking for
  waypoint-based aerial navigation,'' \emph{WSEAS Transactions on Systems and
  Control}, no.~3, pp. 105--119, 2013.

\bibitem{faulwasser2009model}
T.~Faulwasser, B.~Kern, and R.~Findeisen, ``Model predictive path-following for
  constrained nonlinear systems,'' in \emph{Proceedings of the 48h IEEE
  Conference on Decision and Control (CDC) held jointly with 2009 28th Chinese
  Control Conference}.\hskip 1em plus 0.5em minus 0.4em\relax IEEE, 2009, pp.
  8642--8647.

\bibitem{Schwarting_Alonso-Mora_Paull_Karaman_Rus_2018}
W.~Schwarting, J.~Alonso-Mora, L.~Paull, S.~Karaman, and D.~Rus, ``Safe
  nonlinear trajectory generation for parallel autonomy with a dynamic vehicle
  model,'' \emph{IEEE Transactions on Intelligent Transportation Systems},
  vol.~19, no.~9, p. 2994–3008, Sep 2018.

\bibitem{Ji_Zhou_Xu_Gao_2021}
J.~Ji, X.~Zhou, C.~Xu, and F.~Gao, ``Cmpcc: Corridor-based model predictive
  contouring control for aggressive drone flight,'' in \emph{ISER}, 2020.

\bibitem{brito2019model}
B.~Brito, B.~Floor, L.~Ferranti, and J.~Alonso-Mora, ``Model predictive
  contouring control for collision avoidance in unstructured dynamic
  environments,'' \emph{IEEE Robotics and Automation Letters}, vol.~4, no.~4,
  pp. 4459--4466, 2019.

\bibitem{tang2011predictive}
L.~Tang and R.~G. Landers, ``Predictive contour control with adaptive feed
  rate,'' \emph{IEEE/ASME Transactions On Mechatronics}, vol.~17, no.~4, pp.
  669--679, 2011.

\bibitem{Song_Steinweg_Kaufmann_Scaramuzza_2021}
Y.~Song, M.~Steinweg, E.~Kaufmann, and D.~Scaramuzza, ``Autonomous drone racing
  with deep reinforcement learning,'' \emph{2021 IEEE/RSJ International
  Conference on Intelligent Robots and Systems (IROS)}, 2021.

\bibitem{ArcLengthWang2002}
H.~Wang, J.~Kearney, and K.~Atkinson, ``Arc-length parameterized spline curves
  for real-time simulation,'' Jan 2002.

\bibitem{Faessler18ral}
M.~Faessler, A.~Franchi, and D.~Scaramuzza, ``Differential flatness of
  quadrotor dynamics subject to rotor drag for accurate tracking of high-speed
  trajectories,'' \emph{{IEEE} Robot. Autom. Lett.}, Apr. 2018.

\bibitem{waypoint_gaussian_siegwart}
\BIBentryALTinterwordspacing
M.~Neunert, C.~de~Crousaz, F.~Furrer, M.~Kamel, F.~Farshidian, R.~Siegwart, and
  J.~Buchli, ``Fast nonlinear model predictive control for unified trajectory
  optimization and tracking,'' in \emph{2016 IEEE International Conference on
  Robotics and Automation (ICRA)}.\hskip 1em plus 0.5em minus 0.4em\relax IEEE,
  May 2016, p. 1398–1404. [Online]. Available:
  \url{http://ieeexplore.ieee.org/document/7487274/}
\BIBentrySTDinterwordspacing

\bibitem{Mueller13iros}
M.~W. Mueller, M.~Hehn, and R.~D'Andrea, ``A computationally efficient
  algorithm for state-to-state quadrocopter trajectory generation and
  feasibility verification,'' in \emph{IEEE/RSJ Int. Conf. Intell. Robot. Syst.
  (IROS)}, 2013.

\bibitem{bertsekasDP}
D.~P. Bertsekas, \emph{Dynamic Programming and Optimal Control}, 2nd~ed.\hskip
  1em plus 0.5em minus 0.4em\relax Athena Scientific, 2000.

\bibitem{Pfeiffer_Scaramuzza_2021}
C.~Pfeiffer and D.~Scaramuzza, ``Human-piloted drone racing: Visual processing
  and control,'' \emph{IEEE Robotics and Automation Letters}, vol.~6, no.~2, p.
  3467–3474, Apr 2021.

\bibitem{autotune_Loquercio_Saviolo_Scaramuzza_2022}
``Autotune: Controller tuning for high-speed flight,'' vol.~7, p. 4432–4439,
  Apr 2022.

\bibitem{Song_Scaramuzza}
Y.~Song and D.~Scaramuzza, ``Policy search for model predictive control with
  application to agile drone flight,'' \emph{IEEE Transactions on Robotics},
  pp. 1--17, 2022.

\bibitem{HPIPM}
G.~Frison and M.~Diehl, ``\BIBforeignlanguage{en}{Hpipm: a high-performance
  quadratic programming framework for model predictive control},''
  \emph{\BIBforeignlanguage{en}{IFAC-PapersOnLine}}, no.~2, p. 6563–6569, Jan
  2020.

\end{thebibliography}
